\documentclass[11pt]{article}
\usepackage{enumerate}
\usepackage{pdfsync}
\usepackage[OT1]{fontenc}

\usepackage{url}           
\usepackage{booktabs}    
\usepackage{amsfonts}      
\usepackage{nicefrac}      
\usepackage{microtype}    
\usepackage{color}
\usepackage[colorlinks,
            linkcolor=red,
            anchorcolor=blue,
            citecolor=blue, pagebackref=true
           ]{hyperref}
\usepackage{fullpage}
\usepackage{setspace}
\usepackage{tabularx}
\usepackage{float}
\usepackage{wrapfig,lipsum}
\usepackage{enumitem}
\usepackage{natbib}
\usepackage{multirow} 
\usepackage[dvipsnames]{xcolor}
\usepackage{subcaption}
\usepackage{algorithm}
\usepackage{algpseudocode}

\makeatletter
\newcommand*{\rom}[1]{\expandafter\@slowromancap\romannumeral #1@}
\makeatother
\usepackage{amsmath}

\usepackage{mmll}
\usepackage{fancyvrb}
\usepackage{cleveref}
\usepackage{pifont}%

\usepackage{listings}
\usepackage{xcolor, colortbl}

\usepackage[normalem]{ulem}
\usepackage{tcolorbox}
\usepackage{rotating}

\lstdefinestyle{xmlstyle}{
    backgroundcolor=\color{lightgray},
    basicstyle=\ttfamily\small,
    keywordstyle=\color{purple},
    commentstyle=\color{green},
    morekeywords={option,gravity,timestep},
    columns=flexible,
    breaklines=true
}

\let\oldnl\nl
\newcommand{\nonl}{\renewcommand{\nl}{\let\nl\oldnl}}%

\newcommand{\nop}[1]{}

\usepackage{algorithm}
\usepackage{algpseudocode}

\usepackage{xspace}

\newcommand{\algname}{\textsc{CEDGE}\xspace}

\newcommand{\bestSc}[1]{\cellcolor{cyan!35}{#1}}

\newcommand{\secondbestSc}[1]{\cellcolor{cyan!15}{#1}}

\definecolor{softpurple}{RGB}{150, 120, 180}

\definecolor{myviolet}{RGB}{138, 43, 226}  %

\usepackage{listings}
\usepackage{xcolor, colortbl}

\usepackage[normalem]{ulem}
\usepackage{tcolorbox}
\usepackage{rotating}
\usepackage{enumitem}
\usepackage{subcaption}
\usepackage{wrapfig}

\lstdefinestyle{xmlstyle}{
  basicstyle=\ttfamily\scriptsize,
  breaklines=true,
  breakatwhitespace=true,
  columns=flexible,
  keepspaces=true,
  showstringspaces=false,
  frame=none
}

\allowdisplaybreaks

\begin{document}
\title{\huge Cross-Domain Energy-Guided Diffusion Generation for Off-Dynamics Reinforcement Learning}
\author{
   Yu Yang\thanks{ 
   Duke University; email: {\tt
   yu.yang@duke.edu}}  
   ~~~~~~
   Yihong Guo\thanks{ 
   Johns Hopkins University; email: {\tt
   yguo80@jhu.edu
   }} 
   ~~~~~~
   Anqi Liu\thanks{
   Johns Hopkins University; email: {\tt
   aliu.cs@jhu.edu}}  
   ~~~~~~
   Pan Xu\thanks{
   Duke University; email: {\tt
   pan.xu@duke.edu}}   
}
\date{}
\maketitle

\begin{figure}[h]
    \centering
    \includegraphics[width=0.9\linewidth]{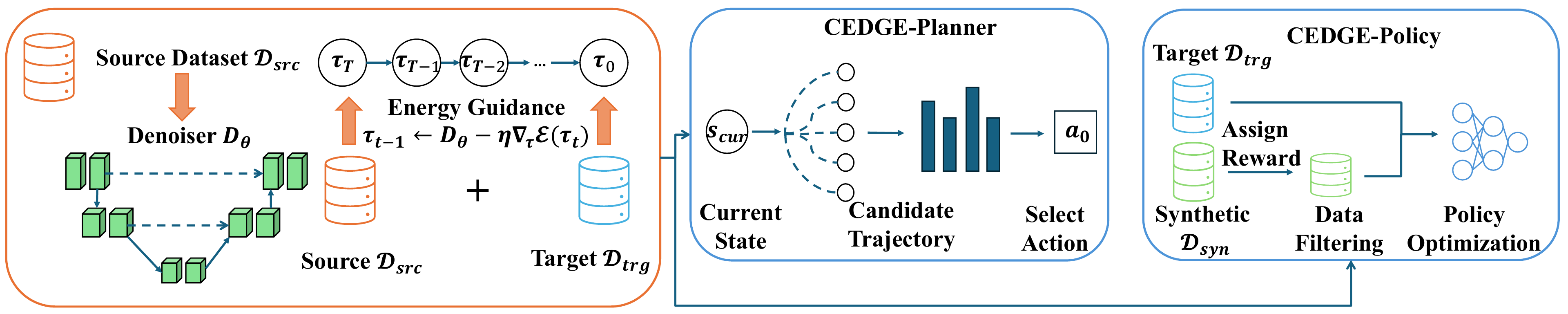}
    \caption{Overview of \algname{}. A source-domain trajectory diffusion model is adapted to target dynamics through learned energy guidance. The resulting guided trajectories can be utilized either for direct planning or as high-quality synthetic data for downstream policy optimization.}
    \label{fig:overview}
\end{figure}

\begin{abstract}
Off-dynamics offline reinforcement learning seeks to learn a target-domain policy from a large source dataset and a limited target dataset under mismatched transition dynamics. Existing approaches such as reward augmentation and data filtering are constrained to the source dataset and cannot synthesize new target behavior to improve coverage beyond the collected source trajectories. While recent model-based methods attempt to address this by learning target-aware dynamics, the generated experience is constructed only at the transition level, which leads to accumulated errors over long horizons. These limitations necessitate a shift toward trajectory-level generation for off-dynamics offline RL. We propose CEDGE, a Cross-domain Energy-guided Diffusion GEneration framework. CEDGE trains a trajectory diffusion model on source-domain trajectories and adapts the generated samples to the target domain through energy guidance. This guidance is derived by minimizing the distribution mismatch between the source and desired target-domain trajectories and is decomposed into return, domain, and behavior energy components. The resulting energy-guided trajectories are useful both for direct planning and as synthetic data for policy learning. Since target adaptation is achieved via energy guidance rather than retraining the diffusion model, CEDGE can be efficiently adapted to new target dynamics compared to previous methods. Experiments on the ODRL benchmark demonstrate that trajectory-level energy-guided generation improves diffusion planning under dynamics shifts and produces synthetic data that improves downstream target policy learning.
\end{abstract}

\section{Introduction}

Off-dynamics offline reinforcement learning (RL) \citep{kaelbling1996reinforcement,eysenbach2020off,liu2024distributionally,guo2024off,wang2026return} considers a data regime where a policy has access to a large source dataset and a limited target dataset under shifted transition dynamics. The source dataset typically contains abundant data collected from the source environment, whose dynamics are different from the target environment. 
In contrast, the target dataset is often too limited to train a robust target domain policy on its own. During training, the agent only has access to the pre-collected source and target datasets with no additional interaction with either environment. Therefore, the primary challenge is how to leverage the abundant source-domain experience to improve target-domain policy learning without introducing detrimental dynamics mismatch.

Existing off-dynamics offline RL methods offer partial solutions to this challenge. Reward augmentation techniques modify source rewards using domain classifiers that estimate the dynamics discrepancy between the source and target domains \citep{eysenbach2020off,liu2022dara,lyu2024cross}. However, these methods rely exclusively on the source dataset throughout policy learning. 
Consequently, source transitions that are unlikely under the target dynamics may still induce harmful transfer and degrade target-domain performance.
To mitigate the influence of mismatched source data, another approach, data filtering, selects source samples that resemble the target domain and discards samples prone to harmful transfer \citep{xu2023cross,wen2024contrastive,lyu2025cross,xia2026localized}. While these methods can reduce harmful transfer from mismatched source data, both approaches are fundamentally limited to the trajectories that have already been collected in the source dataset. Specifically, reward augmentation only augments the rewards of existing source samples, and data filtering only selects from already collected trajectories. Therefore, neither method can synthesize novel target-domain behavior to improve data coverage beyond the inherent support of the collected source trajectories.

Recently, a model-based method, MOBODY, addresses this limitation by learning target-aware dynamics models and generating synthetic rollouts for policy learning \citep{guo2026mobody}. By doing so, policy optimization can extend beyond the initial offline dataset via interaction with the target-aware dynamics model. Despite this advancement, the generated rollouts are still constructed at the transition level. As the rollout horizon increases, even minor one-step prediction errors can compound, causing later states to drift away from the true target-domain state distribution. This accumulation of errors can corrupt the generated trajectory and mislead policy optimization. Consequently, transition-level generation remains inadequate for long-horizon off-dynamics adaptation, which motivates our focus on trajectory-level generation.

Diffusion models \citep{ho2020denoising,song2021denoising,song2021scorebased} offer a natural framework for solving this trajectory-level generation problem in offline RL. Previous work has established that diffusion models can model distributions over complete state-action trajectories and are effective for planning \citep{janner2022planning,wang2022diffusion,ajay2023is}. Furthermore, other studies demonstrate that diffusion-generated trajectories can serve as synthetic data for downstream policy learning \citep{lu2023synthetic,jackson2024policy}. These results suggest that trajectory diffusion models can function not only as test-time planners but also as generators of extra offline trajectories, making this generative perspective highly relevant to off-dynamics offline RL. As discussed above, reward augmentation and data filtering are limited to the source dataset, while transition-level model rollouts may suffer from compounding errors. Diffusion modeling over trajectory distributions offers a distinct solution by generating complete trajectories that potentially expand the training data at the trajectory level. However, merely training a trajectory diffusion model on source-domain trajectories will not work, as the generated samples retain the source transition dynamics and behavior patterns. Conversely, training a robust trajectory diffusion model solely on the limited target dataset is challenging due to insufficient target-domain coverage. Therefore, the central problem we address is how to utilize abundant source-domain trajectories for diffusion generation while simultaneously adapting the generated samples to the target domain.

In this paper, we propose \algname, a cross-domain energy-guided diffusion generation framework for off-dynamics offline RL. Our method first trains a reusable trajectory diffusion model using source-domain trajectories and then employs energy-guided sampling to steer the generated samples toward the target domain. The resulting energy-guided trajectories facilitate two downstream applications: direct diffusion planning and offline policy optimization. We formally derive the energy guidance from the distributional mismatch between the source trajectory distribution and the desired target-domain trajectory distribution. This guidance decomposes into return, domain, and behavior energy terms, which encourages high-return trajectories, target-domain dynamic compatibility, and effective compensation for source behavior bias, respectively. This mechanism allows our method to generate trajectories more compatible with the target domain, rather than simply reusing source data or constructing long-horizon rollouts from a transition-level model. By guiding complete trajectories during the diffusion sampling process, we effectively mitigate the accumulated errors inherent in transition-level generation. Moreover, because target adaptation is achieved through energy guidance rather than diffusion model retraining, CEDGE allows for fast adaptation to new target dynamics. The overall architecture is summarized in \Cref{fig:overview}.

\section{Related Work}
\label{sec:related}
\paragraph{Diffusion models in RL.}
Diffusion models \citep{ho2020denoising,karras2022elucidating} have recently become a useful tool for decision-making and offline reinforcement learning. A significant line of work utilizes diffusion models for trajectory-level planning. For example, Diffuser \citep{janner2022planning} formulates planning as trajectory generation, showing that diffusion models can model long-horizon trajectory structure and support flexible trajectory-level planning. More recently, Veteran \citep{lu2025makes} provides systematic analyses of diffusion planning, identifying which architectural and sampling choices are more appropriate for achieving high performance on offline RL tasks. Another line of work uses diffusion models as expressive policy classes for offline RL. Diffusion-QL \citep{wang2022diffusion} and IDQL \citep{hansen2023idql} use diffusion-based action modeling combined with value-guided policy improvement, demonstrating that diffusion policies can represent complex and multimodal action distributions. Beyond direct planning or policy representation, recent studies \citep{lu2023synthetic,jackson2024policy} have also explored using diffusion models as data generators for downstream policy learning. In particular, PGD \citep{jackson2024policy} uses diffusion models to synthesize policy-guided trajectories for offline RL, highlighting the potential of generated trajectories as synthetic training data rather than only for test-time planning. Compared to these works, our method leverages diffusion not just as a planner or a policy class, but as a source-domain trajectory prior that can be adapted to novel target dynamics through energy guidance and subsequently coupled with offline policy optimization.

\paragraph{Energy-guided diffusion in RL.}
Energy-guided diffusion provides a flexible mechanism for steering a learned diffusion model toward a desired distribution without retraining the generator. In offline RL, CEP \citep{lu2023contrastive} studies energy-guided diffusion sampling theoretically and proposes contrastive energy prediction to estimate the intermediate guidance required for exact energy-guided sampling. EDIS \citep{liu2024energy} learns a diffusion model from offline data and uses energy guidance to generate samples that better match the distribution induced by the policy during online fine-tuning. EDIS shows that energy-guided diffusion can mitigate the mismatch between fixed offline data and the distribution encountered during online learning. However, EDIS focuses on the offline-to-online setting, where additional online interaction is available and the policy is explicitly improved through fine-tuning. In contrast, we study fully offline off-dynamics RL, where no additional target-domain interaction is allowed and the central challenge is the transition dynamics mismatch between source and target domains.

\paragraph{Off-dynamics RL.}
Off-dynamics reinforcement learning \citep{eysenbach2020off,liu2022dara,liu2024upper,tang2024robust,he2025sample,NEURIPS2025_855b928e} addresses the challenge of transferring policies across domains with mismatched transition dynamics. A common approach is reward augmentation based on domain classifiers. For instance, DARC \citep{eysenbach2020off} derives reward augmentation from the source-target dynamics discrepancy, and DARA \citep{liu2022dara} extends this idea to the offline setting. Another prominent line of work focuses on selecting or filtering source data that aligns more closely with the target domain. BOSA \citep{liu2024beyond} proposes off-dynamics offline RL under dataset support constraints. Data filtering methods utilize contrastive representations \citep{wen2024contrastive}, optimal transport \citep{lyu2025cross}, or localized dynamics-aware filtering \citep{xia2026localized} to select data from the source dataset. While these methods effectively mitigate harmful transfer from mismatched source data, their major limitation is that they remain largely restricted to the original collected offline dataset. More recently, a model-based approach, MOBODY \citep{guo2026mobody}, attempts to surpass the dataset support constraint by learning target-aware dynamics and generating synthetic transitions for policy learning.
Our work differentiates itself by addressing off-dynamics offline RL at the trajectory level. Instead of correcting individual transitions or learning a comprehensive target dynamics model for rollouts, we adapt a source-domain diffusion trajectory model to the target domain using energy-guided sampling. We then utilize the resulting generated trajectories for downstream offline policy optimization.

\section{Preliminary}

\paragraph{Off-dynamics Offline RL}
We begin by defining reinforcement learning within the standard Markov Decision Process (MDP) framework, characterized by the tuple $\mathcal{M} = (\mathcal{S}, \mathcal{A}, \mathcal{R}, p, \gamma, \rho_0)$. Here, $\mathcal{S}$ and $\mathcal{A}$ denote the state and action spaces, respectively; $\mathcal{R}: \mathcal{S} \times \mathcal{A} \rightarrow \mathbb{R}$ is the reward function; $p(s' \mid s,a)$ defines the transition dynamics; $\gamma \in (0,1)$ is the discount factor; and $\rho_0$ is the initial-state distribution. In the off-dynamics setting, we consider two distinct operational domains: a source domain $\mathcal{M}_{\mathrm{src}} = (\mathcal{S}, \mathcal{A}, \mathcal{R}, p_{\mathrm{src}}, \gamma, \rho_0)$ and a target domain $\mathcal{M}_{\mathrm{trg}} = (\mathcal{S}, \mathcal{A}, \mathcal{R}, p_{\mathrm{trg}}, \gamma, \rho_0)$. These two environments share the same state space, action space, reward function, and discount factor, but crucially differ in their underlying transition dynamics, such that $p_{\mathrm{src}}(s' \mid s,a) \neq p_{\mathrm{trg}}(s' \mid s,a)$.

Our work specifically focuses on the challenging paradigm of off-dynamics offline RL. We assume access to a large, pre-collected offline dataset from the source domain, $\mathcal{D}_{\mathrm{src}}$, and a limited offline dataset from the target domain, $\mathcal{D}_{\mathrm{trg}}$. Both datasets were collected by corresponding behavior policies in their respective environments, and importantly, no further environmental interaction is permitted during the training process. The objective is to learn a policy $\pi$ that maximizes the expected discounted return within the target domain:
$
J_{\mathrm{trg}}(\pi)
=
\mathbb{E}_{\pi, p_{\mathrm{trg}}, \rho_0}
\left[
\sum_{t=0}^{\infty} \gamma^t r(s_t,a_t)
\right].
$

\paragraph{Generative Modeling via Diffusion Models}
Diffusion models learn a data distribution through a two-stage framework: a forward noising process and a reverse denoising process \citep{ho2020denoising}. The forward process progressively perturbs the input data with Gaussian noise, while the reverse process is trained to iteratively remove this noise, ultimately recovering a sample from the original data distribution. In our setting, the data sample is defined as a trajectory segment. Let $\tau_0 = (s_0,a_0,\ldots,s_{H-1},a_{H-1},s_H)$ denote a trajectory segment of horizon $H$. We model the diffusion process across $T$ denoising steps. Let $\{\beta_k\}_{k=1}^{T}$ denote the variance schedule, and define $\alpha_k=1-\beta_k$ and $\bar{\alpha}_k=\prod_{i=1}^{k}\alpha_i$.
For each step $k\in\{1,\ldots,T\}$, the forward noising process is defined as
$
q(\tau_k \mid \tau_0) =
\mathcal{N}(
\tau_k;\,
\sqrt{\bar{\alpha}_k}\,\tau_0,\,
(1-\bar{\alpha}_k)I
),
$
Following existing diffusion planning methods \citep{janner2022planning,lu2025makes}, we train a denoiser $D_\theta(\tau_k,k)$ on the source-domain trajectories ($\mathcal{D}_{\mathrm{src}}$) to predict the clean trajectory $\tau_0$ from a noisy sample $\tau_k$. The training objective is:
\begin{align}
\label{eq:diffusion_loss}
\mathcal{L}_{\mathrm{diff}}(\theta)=
\mathbb{E}_{\tau_0 \sim \mathcal{D}_{\mathrm{src}}, k, \epsilon \sim \mathcal{N}(0,I)}
\big[\|D_\theta(\tau_k,k)-\tau_0 \|_2^2 \big],
\end{align}
where $\tau_k = \sqrt{\bar{\alpha}_k} \tau_0 + \sqrt{1-\bar{\alpha}_k}\epsilon$. After training, this denoiser defines a trajectory prior over the source-domain data distribution. 
Energy-guided diffusion steers the learned trajectory generator toward a desired distribution by reweighting its generated trajectory distribution with an energy function. Let $q_\theta(\tau)$ denote the trajectory distribution induced by the learned diffusion model. Given an energy function $\mathcal E(\tau)$, the guided trajectory distribution can be written as
\begin{align}
p(\tau)
\propto
q_\theta(\tau)\exp\left(-\mathcal E(\tau)\right),
\label{eq:energy_guided_distribution}
\end{align}
where trajectories with lower energy are assigned higher probability. This mechanism has also been studied in prior work on energy-guided diffusion sampling \citep{lu2023contrastive,chung2023diffusion}.

\section{Cross-Domain Energy-Guided Diffusion Generation}
\label{sec:method}

In this section, we introduce \algname, a cross-domain energy-guided diffusion generation framework specifically designed to address the challenges of off-dynamics offline reinforcement learning. We first derive the trajectory-level energy guidance necessary to adapt a source-domain diffusion prior to the target dynamics. Subsequently, we detail how the three distinct energy modules are learned, and lastly, we explain how the guided sampler is employed for both policy planning and synthetic-data generation.

\subsection{Trajectory-Level Energy Guidance}
\label{sec:trajectory_energy_guidance}
We first formally define the energy guidance mechanism to adapt generated trajectories to the target domain. Given the limited availability of target-domain data ($\mathcal{D}_{\mathrm{trg}}$), training a reliable trajectory diffusion model directly on $\mathcal{D}_{\mathrm{trg}}$ is generally challenging. \algname trains its trajectory diffusion model on the larger source-domain dataset ($\mathcal{D}_{\mathrm{src}}$) and subsequently adapts the generated samples to the target domain using energy-guided sampling. The core idea is to design this energy guidance such that the generated trajectories align with the target dynamics and achieve high expected returns in the target environment. To this end, we derive the energy guidance by examining the distributional mismatch between the source trajectory distribution and a desired target-domain trajectory distribution.

Let $\tau=(s_0,a_0,\ldots,s_{H-1},a_{H-1},s_H)$ be a trajectory segment of horizon $H$. We denote the desired target-domain trajectory distribution by $p^\star(\tau)$. Drawing upon the probabilistic inference perspective in RL \citep{levine2018reinforcement}, the desired target distribution is defined as:
\begin{align}
\label{eq:target_distribution}
p^{\star}(\tau)
\propto
\rho_0(s_0)
\big(\textstyle\prod_{t=0}^{H-1} p_{\mathrm{trg}}(s_{t+1}\mid s_t,a_t)\big)
\exp\big(\textstyle\sum_{t=0}^{H-1} \gamma^t r(s_t,a_t)\big),
\end{align}
where $\rho_0(s_0)$ is the initial-state distribution, $p_{\mathrm{trg}}(s_{t+1}\mid s_t,a_t)$ is the target-domain transition dynamics, and the exponential return term weights trajectories proportionally to their discounted return. Crucially, this distribution represents the {\it ideal} target trajectories, not merely the empirical distribution of the limited $\mathcal{D}_{\mathrm{trg}}$. Thus, energy-guided trajectories must simultaneously adhere to target dynamics and maximize returns in the target environment.

The source trajectory distribution, $q_{\mathrm{src}}(\tau)$, is induced by the source domain transition dynamics $p_{\mathrm{src}}(s_{t+1}\mid s_t,a_t)$ and the behavior policy $\pi_\beta(a_t\mid s_t)$ that generated the source dataset:
\begin{align}
\label{eq:source_distribution}
q_{\mathrm{src}}(\tau) =\rho_0(s_0)
\big(\textstyle\prod_{t=0}^{H-1} p_{\mathrm{src}}(s_{t+1}\mid s_t,a_t)
\pi_\beta(a_t\mid s_t)\big).
\end{align}
We assume a common initial-state distribution, $\rho_0$, for both domains. Since the trajectory diffusion model is trained on $\mathcal{D}_{\mathrm{src}}$, it inherently approximates $q_{\mathrm{src}}(\tau)$. To adapt the samples generated by this model to the target domain, we define the energy-guided trajectory distribution as:
$
p^{\star}(\tau)
\propto
q_{\mathrm{src}}(\tau)\exp\bigl(-\mathcal{E}(\tau)\bigr),
$
where $\mathcal{E}(\tau)$ is the energy function utilized during the reverse diffusion sampling process. The energy guidance reweights the source trajectory distribution toward the desired target-domain distribution. Trajectories with lower energy $\mathcal{E}(\tau)$ are consequently assigned a higher probability after guidance. Equivalently, the energy function can be expressed as:
$
\label{eq:energy_density_ratio}
\mathcal{E}(\tau)
=\log q_{\mathrm{src}}(\tau)
-\log p^\star(\tau)+C,
$
where $C$ is a constant independent of $\tau$ and does not affect gradient-based guidance. Then we can get 
{\small
\begin{align*}
\mathcal{E}(\tau)
= \textstyle\sum_{t=0}^{H-1}
[ \log p_{\mathrm{src}}(s_{t+1}\mid s_t,a_t) -\log p_{\mathrm{trg}}(s_{t+1}\mid s_t,a_t)-\gamma^t r(s_t,a_t)+\log \pi_\beta(a_t\mid s_t)]+C.
\end{align*}}%
This shows that the energy guidance comprises three distinct, additive terms: domain correction ($\mathcal{E}_1$), return maximization ($\mathcal{E}_2$), and behavior policy compensation ($\mathcal{E}_3$), defined as:
$\mathcal{E}_{1}(\tau) =\textstyle \sum_{t=0}^{H-1}[\log p_{\mathrm{src}}(s_{t+1}\mid s_t,a_t)-\log p_{\mathrm{trg}}(s_{t+1}\mid s_t,a_t)]+C_{1}$,
$\mathcal{E}_{2}(\tau) =\textstyle -\sum_{t=0}^{H-1}\gamma^t r(s_t,a_t)+C_{2}$ and 
$\mathcal{E}_{3}(\tau) =\textstyle \sum_{t=0}^{H-1}\log \pi_\beta(a_t\mid s_t)+C_{3}$.
More specifically, $\mathcal{E}_{1}$ is the domain energy guidance. 
It assigns lower energy to transitions more likely under the target-domain dynamics than under the source-domain dynamics. 
$\mathcal{E}_{2}$ is the return energy guidance. Since it is the negative discounted return, minimizing this term favors trajectories with higher returns in the target environment. 
$\mathcal{E}_{3}$ is the policy energy guidance. 
It compensates for the source behavior policy bias present in $q_{\mathrm{src}}(\tau)$, thereby reducing the action preference bias inherited from the source-domain offline data. 
These terms are combined in practice using weighted coefficients $\lambda_{i}$ during reverse diffusion:
\begin{align}
\label{eq:weighted_energy}
\mathcal{E}(\tau)
=
\lambda_{1}\mathcal{E}_{1}(\tau)
+
\lambda_{2}\mathcal{E}_{2}(\tau)
+
\lambda_{3}\mathcal{E}_{3}(\tau).
\end{align}

\subsection{Learning Energy Guidance Functions}
We now detail how each guidance term in \eqref{eq:weighted_energy} is estimated from the available offline data. 

\paragraph{Domain energy function.} 
To implement $\mathcal{E}_{1}(\tau)$, which quantifies the transition dynamics mismatch, we adapt prior work \citep{eysenbach2020off,liu2022dara} and utilize two domain classifiers. We can formulate the directional gradient of the energy guidance as:
\begin{align*}  
    \nabla_{\tau} \mathcal{E}_{1}(\tau) & = -\nabla_{\tau}\textstyle\sum_{t=0}^{H-1}\big[\log p(\text{trg} \mid s_t, a_t, s_{t+1})-\log p(\text{trg} \mid s_t, a_t)\\ \notag
    & \qquad-\log p(\text{src} \mid s_t, a_t, s_{t+1})+\log p(\text{src} \mid s_t, a_t)\big]. 
\end{align*}
This gradient utilizes two domain classifiers: $p_{\theta_{\mathrm{SAS}}}$ classifies the transition $(s_t, a_t, s_{t+1})$ as target or source, and $p_{\theta_{\mathrm{SA}}}$ classifies the state-action pair $(s_t, a_t)$ as target or source. The classifiers are trained by minimizing the following cross-entropy loss functions:
\begin{align}
\mathcal{L}_{\mathrm{SAS}}
&= -\mathbb{E}_{\mathcal{D}_{\mathrm{src}}}
      \bigl[\log p_{\theta_{\mathrm{SAS}}}(\text{src}\mid s,a,s')\bigr]
   - \mathbb{E}_{\mathcal{D}_{\mathrm{trg}}}
      \bigl[\log p_{\theta_{\mathrm{SAS}}}(\text{trg}\mid s,a,s')\bigr], \label{eq:loss_sas}\\
\mathcal{L}_{\mathrm{SA}}
&= -\mathbb{E}_{\mathcal{D}_{\mathrm{src}}}
      \bigl[\log p_{\theta_{\mathrm{SA}}}(\text{src}\mid s,a)\bigr]
   - \mathbb{E}_{\mathcal{D}_{\mathrm{trg}}}
      \bigl[\log p_{\theta_{\mathrm{SA}}}(\text{trg}\mid s,a)\bigr]. \label{eq:loss_sa}
\end{align}
where $\mathcal{D}_{\mathrm{src}}$ and $\mathcal{D}_{\mathrm{trg}}$ denote the source and target datasets, respectively.

\paragraph{Return energy function.}
Next, we implement the return guidance term $\mathcal{E}_{2}(\tau)$, which steers the trajectory generation towards high-return trajectories. Since $\mathcal{E}_{2}(\tau)$ must assign lower energy to trajectories with larger returns, we train a differentiable return predictor $J_{\psi}(\tau)$ that estimates the discounted cumulative reward for a trajectory segment $\tau$. We train this predictor on the combined trajectory dataset $\mathcal{D}_{\mathrm{src}} \cup \mathcal{D}_{\mathrm{trg}}$  by minimizing the standard regression loss:
\begin{align}
\label{eq:return_energy}
\mathcal L_{R}
=
\mathbb E_{\tau \sim \mathcal D_{\mathrm{src}} \cup \mathcal D_{\mathrm{trg}}}
\big\|
J_{\psi}(\tau) - \textstyle\sum_{t=0}^{H-1}\gamma^t r(s_t,a_t)
\big\|_2^2.
\end{align}
The energy guidance is then derived from the predictor's gradient $ \nabla_{\tau}\mathcal E_{2}(\tau) = - \nabla_{\tau} J_{\psi}(\tau).$
This ensures that the reverse diffusion process is guided towards trajectories with higher predicted return. 

\paragraph{Policy energy function.}
Lastly, we address the third guidance term, $\mathcal{E}_{3}(\tau)$, which compensates for the source behavior policy bias embedded in the source diffusion model. %
Since the true source behavior policy $\pi_\beta$ is unknown, we approximate it by learning a Gaussian policy model, $\pi_{\omega}(a \mid s)$, via behavior cloning on $\mathcal{D}_{\mathrm{src}}$. We train $\pi_{\omega}$ by maximizing the likelihood of source actions, equivalent to minimizing the negative log-likelihood:
\begin{align}
\label{eq:policy_energy}
\mathcal L_{\pi}
= -\mathbb E_{(s,a)\sim \mathcal D_{\mathrm{src}}}
\big[ \log \pi_{\omega}(a \mid s) \big].
\end{align}
Using the learned Gaussian behavior policy, we define the policy energy guidance as
$
\mathcal E_{3}(\tau) =\sum_{t=0}^{H-1}\log \pi_{\omega}(a_t \mid s_t),
$
and its gradient with respect to the trajectory is given by
$\nabla_{\tau}\mathcal E_{3}(\tau) =\nabla_{\tau}\sum_{t=0}^{H-1}\log \pi_{\omega}(a_t \mid s_t).$ This gradient is applied during reverse diffusion to mitigate the bias toward source domain action preferences inherited from the offline data.

\paragraph{Training and inference protocol.} 
The process is conducted sequentially. First, the trajectory diffusion model is trained exclusively on source-domain trajectories (as detailed in \Cref{alg:diffusion_training} in \Cref{app:technical_details}). This initial step establishes a robust generative model for the source trajectory distribution without requiring any target-domain data. Subsequently, for each new target environment, the trained denoiser is fixed, and the energy guidance functions ($\mathcal{E}_1, \mathcal{E}_2, \mathcal{E}_3$) are trained using both the source and target offline datasets. Specifically, the domain classifier estimates the dynamics mismatch (corresponding to $\mathcal{E}_1$), the return predictor estimates the return guidance (corresponding to $\mathcal{E}_2$), and the Gaussian policy estimates the behavior-policy factor (corresponding to $\mathcal{E}_3$). Finally, during the inference phase (detailed in \Cref{alg:energy_guided_sampling} in \Cref{app:technical_details}), the learned energy functions guide the reverse diffusion process. At every reverse diffusion step, the denoiser first predicts a clean trajectory estimate, which is then transformed into a diffusion score. This score is subsequently augmented by the energy guidance score to generate an energy-guided reverse update.

\subsection{Diffusion as the Planner}
The trained source trajectory diffusion model can be directly employed as a planner during inference time \citep{janner2022planning,lu2025makes}. The key idea is to generate energy-guided trajectory candidates conditioned on the current state, and select the trajectory with the highest predicted return for action execution.

Given the current state $s_{\mathrm{cur}}$, we generate trajectory candidates of horizon $H$ by running the energy-guided reverse diffusion process conditioned on $s_{\mathrm{cur}}$. Specifically, we model the current-state-conditioned target trajectory distribution as:
$
p^{\star}(\tau \mid s_0 = s_{\mathrm{cur}})
\propto
q_{\mathrm{src}}(\tau \mid s_0 = s_{\mathrm{cur}})
\exp\bigl(-\mathcal E(\tau)\bigr).
$
Starting from Gaussian noise, we iteratively denoise the trajectory using the learned denoiser combined with the guidance gradients derived from the three energy functions. Crucially, the first state of the generated trajectory is fixed to the current state $s_{\mathrm{cur}}$, ensuring that every sampled trajectory is accurately anchored at the current decision point.

At each decision step, multiple trajectory candidates are generated by repeating the guided reverse diffusion process. For each generated trajectory $\hat\tau$, we evaluate its quality using the learned return energy $\mathcal E_2(\hat\tau)$. Since the return energy is defined as the negative predicted return, minimizing $\mathcal E_2$ is equivalent to selecting the trajectory with the highest predicted return. We then select the trajectory with the lowest return energy, $\hat\tau^{\star} = \arg\min_{\hat\tau} \mathcal E_2(\hat\tau)$, and execute its first action. Following the observation of the resulting next state, we repeat this entire procedure from the new current state.

\subsection{Policy Optimization on Generated Data}
While the diffusion planner serves as a direct inference-time decision maker, we can also utilize the generated trajectories to augment the offline data for better policy optimization.

\paragraph{Reward annotation.}
First, we employ the learned diffusion model along with the guidance modules to synthesize a set of target trajectories. Although the environment reward is defined as $r(s,a)$, we parameterize a reward model as $\hat r_\eta(s,a,s')$ to allow it to leverage transition information when annotating the generated samples. $\hat r_\eta$ is trained on both the source and target datasets by minimizing:
\begin{align}
\label{eq:reward_loss}
\mathcal L_{\mathrm{reward}}
= \mathbb E_{(s,a,s')\sim \mathcal D_{\mathrm{src}} \cup \mathcal D_{\mathrm{trg}}}
[
\hat r_\eta(s,a,s') - r(s,a)
]^2.
\end{align}
Then $\hat r_\eta(s,a,s')$ is used to assign rewards to transitions extracted from the generated trajectories.

\paragraph{Trajectory-level quality control and policy optimization.}
Generated trajectories may still contain samples that exhibit poor alignment with the target dynamics or possess low predicted returns. Therefore, before constructing the synthetic dataset, we apply a rigorous two-stage trajectory-level filtering procedure. In the first stage, we utilize the dynamics energy $\mathcal E_1(\hat\tau)$ to quantify the alignment with the target-domain dynamics, retaining only trajectories with lower dynamics energy. In the second stage, we rank the remaining trajectories based on their return energy $\mathcal E_2(\hat\tau)$ and keep only those with lower return energy. Let $\mathcal D_{\mathrm{syn}}$ denote this filtered synthetic dataset. We then train the final target-domain policy using IQL \citep{kostrikov2021offline} on the combined dataset $\mathcal D_{\mathrm{syn}} \cup \mathcal D_{\mathrm{trg}}$.

\section{Experiments}
In this section, we conduct experiments using the benchmark dataset provided by ODRL \citep{lyu2024odrlbenchmarkoffdynamicsreinforcement} to rigorously evaluate the performance of \algname in solving off-dynamics offline RL problems. We evaluate \algname across multiple MuJoCo environments under various levels of dynamics shift.

\subsection{Experimental Setup}

\paragraph{Task and Environments.}
We evaluate \algname on MuJoCo environments with diverse dynamics shifts, following the protocol established in the ODRL benchmark \citep{lyu2024odrlbenchmarkoffdynamicsreinforcement}. We focus on two primary types of dynamics shifts: gravity and friction. For each shift type, we rescaled the corresponding environment parameter by factors of $\{0.1, 0.5, 2.0, 5.0\}$ to introduce a range of dynamics mismatches. The source domain remains the standard, unmodified MuJoCo environment and is shared across all target domains. The target domains are created by applying the specified gravity or friction rescaling factors. This setup matches the fast-adaptation setting targeted by \algname. Since the source domain trajectory diffusion model can be pre-trained once on the source dataset and reused across multiple target domains, only the target-specific energy guidance functions need to be learned for each shifted target environment. We use the dataset in ODRL \citep{lyu2024odrlbenchmarkoffdynamicsreinforcement}. More details are available in \Cref{app:env_details}.

\paragraph{Baselines.} 
We compare our method against both standard offline RL baselines and specialized off-dynamics offline RL baselines. For standard offline RL, we include IQL \citep{kostrikov2021offline} and TD3-BC \citep{fujimoto2021minimalist}. For off-dynamics offline RL, we incorporate DARA \citep{liu2022dara}, BOSA \citep{liu2024beyond}, and MOBODY \citep{guo2026mobody}. We also include a general Diffusion planning baseline \citep{janner2022planning}. Comprehensive implementation details and hyperparameter configurations for all compared baselines are provided in \Cref{app:baseline_details}.

\paragraph{Implementation Details.}
We implement \algname as an energy-guided trajectory diffusion model operating on normalized state-action sequences. The source domain trajectory diffusion model is parameterized by a temporal 1D U-Net denoiser and is trained solely on source trajectories. Energy guidance is implemented through three distinct learned energy modules: domain energy ($\mathcal{E}_1$), return energy ($\mathcal{E}_2$), and policy energy ($\mathcal{E}_3$). The weighting coefficients for the return and domain guidance are selected from the set $\{0.1, 0.5, 1.0, 2.0\}$, while the policy guidance weight is fixed at $0.1$. Additional architecture and hyperparameter specifics are detailed in \Cref{app:implement_details}.

\begin{table}[ht]
\centering
\caption{Performance of our methods and baselines on MuJoCo tasks (HalfCheetah, Ant, Walker2d, Hopper) under medium-level offline datasets with dynamics shifts in gravity and friction at levels $\{0.1, 0.5, 2.0, 5.0\}$. Source domains remain unchanged; target domains are shifted. We report normalized target-domain scores (mean $\pm$ std over three seeds). Best and second-best mean scores are marked in each row.}
\label{tab:mujoco_friction_gravity}
\resizebox{\textwidth}{!}{
\begin{tabular}{c|c|cccccc|cc}
\toprule
Env & Level & BOSA & IQL & TD3-BC & DARA & MOBODY & Diffuser 
& \shortstack{\algname{}-Planner} 
& \shortstack{\algname{}-Policy}\\
\midrule
\multirow{4}{*}{\shortstack{HalfCheetah\\Gravity}}
& 0.1 & $9.31\pm1.94$ & $9.62\pm4.27$ & $6.90\pm0.34$ & $12.90\pm1.01$ & \secondbestSc{$14.18\pm1.06$} & $11.19\pm1.18$ & $10.60\pm1.84$ & \bestSc{$\mathbf{20.04}\pm\mathbf{3.52}$} \\
& 0.5 & $43.96\pm5.68$ & $44.23\pm2.93$ & $6.38\pm3.91$ & $46.11\pm1.93$ & $47.18\pm1.23$ & \secondbestSc{$48.45\pm0.68$} & \bestSc{$\mathbf{49.84}\pm\mathbf{0.17}$} & $2.54\pm0.59$ \\
& 2.0 & $27.86\pm0.94$ & $31.34\pm1.68$ & $29.29\pm3.62$ & $31.85\pm1.31$ & \secondbestSc{$41.60\pm7.35$} & $27.38\pm0.05$ & $27.19\pm0.06$ & \bestSc{$\mathbf{41.66}\pm\mathbf{5.84}$} \\
& 5.0 & $17.95\pm11.97$ & $44.00\pm23.13$ & $73.75\pm14.11$ & $27.67\pm17.01$ & \bestSc{$\mathbf{83.05}\pm\mathbf{1.21}$} & $20.33\pm3.22$ & $4.04\pm0.07$ & \secondbestSc{$75.34\pm1.63$} \\
\midrule

\multirow{4}{*}{\shortstack{HalfCheetah\\Friction}}
& 0.1 & $12.53\pm3.61$ & $26.39\pm11.35$ & $8.95\pm0.71$ & $23.69\pm16.46$ & \secondbestSc{$57.53\pm2.49$} & $10.70\pm0.48$ & $10.55\pm0.05$ & \bestSc{$\mathbf{64.31}\pm\mathbf{0.39}$} \\
& 0.5 & $68.93\pm0.35$ & $69.80\pm0.64$ & $49.43\pm9.91$ & $64.89\pm3.04$ & $69.54\pm0.48$ & \secondbestSc{$71.25\pm0.39$} & \bestSc{$\mathbf{72.13}\pm\mathbf{0.27}$} & $52.06\pm6.41$ \\
& 2.0 & $46.53\pm0.37$ & $46.04\pm2.04$ & $43.51\pm0.74$ & $46.25\pm2.36$ & \bestSc{$\mathbf{50.02}\pm\mathbf{3.26}$} & $46.91\pm0.62$ & $47.93\pm0.14$ & \secondbestSc{$49.15\pm4.03$} \\
& 5.0 & $44.07\pm9.07$ & $44.96\pm6.78$ & $35.83\pm6.65$ & $40.06\pm7.87$ & \secondbestSc{$59.20\pm4.91$} & $32.11\pm1.80$ & $32.79\pm0.91$ & \bestSc{$\mathbf{68.56}\pm\mathbf{1.94}$} \\
\midrule

\multirow{4}{*}{\shortstack{Ant\\Gravity}}
& 0.1 & $25.58\pm2.21$ & $12.53\pm1.11$ & $13.23\pm2.61$ & $11.03\pm1.24$ & \secondbestSc{$37.09\pm2.12$} & $11.92\pm1.57$ & $11.12\pm0.11$ & \bestSc{$\mathbf{37.99}\pm\mathbf{1.13}$} \\
& 0.5 & $19.03\pm4.41$ & $10.09\pm2.00$ & $12.91\pm2.85$ & $9.04\pm1.35$ & \bestSc{$\mathbf{37.44}\pm\mathbf{2.79}$} & $12.44\pm2.01$ & $7.13\pm0.24$ & \secondbestSc{$35.16\pm4.51$} \\
& 2.0 & $41.77\pm1.52$ & $37.17\pm0.96$ & $34.04\pm4.12$ & $36.64\pm0.82$ & \bestSc{$\mathbf{45.83}\pm\mathbf{1.71}$} & $35.84\pm1.32$ & $34.11\pm0.96$ & \secondbestSc{$44.42\pm1.26$} \\
& 5.0 & $31.94\pm0.69$ & $31.59\pm0.35$ & $6.37\pm0.45$ & $31.01\pm0.39$ & \secondbestSc{$65.45\pm3.23$} & $10.46\pm1.08$ & $27.42\pm0.12$ & \bestSc{$\mathbf{71.46}\pm\mathbf{6.56}$} \\
\midrule

\multirow{4}{*}{\shortstack{Ant\\Friction}}
& 0.1 & \secondbestSc{$58.95\pm0.71$} & $55.56\pm0.46$ & $49.20\pm2.55$ & $55.12\pm0.24$ & $58.79\pm0.11$ & $57.07\pm3.41$ & $48.41\pm0.31$ & \bestSc{$\mathbf{61.85}\pm\mathbf{3.94}$} \\
& 0.5 & \secondbestSc{$59.72\pm3.57$} & $59.28\pm0.80$ & $25.21\pm7.17$ & $58.92\pm0.80$ & \bestSc{$\mathbf{62.41}\pm\mathbf{4.10}$} & $58.98\pm4.19$ & $42.36\pm0.56$ & $54.06\pm1.93$ \\
& 2.0 & $20.18\pm3.79$ & $19.84\pm3.20$ & $22.69\pm8.10$ & $17.54\pm2.47$ & \secondbestSc{$47.41\pm4.40$} & $27.41\pm2.31$ & $10.66\pm0.49$ & \bestSc{$\mathbf{57.49}\pm\mathbf{4.38}$} \\
& 5.0 & $9.07\pm0.88$ & $7.75\pm0.25$ & $10.06\pm4.16$ & $7.80\pm0.12$ & \secondbestSc{$31.17\pm5.57$} & $15.67\pm5.24$ & $12.48\pm0.99$ & \bestSc{$\mathbf{60.31}\pm\mathbf{9.32}$} \\
\midrule

\multirow{4}{*}{\shortstack{Walker2d\\Gravity}}
& 0.1 & $18.75\pm12.02$ & $16.04\pm7.60$ & $36.48\pm0.95$ & $20.12\pm5.74$ & \secondbestSc{$65.85\pm5.08$} & $19.78\pm0.73$ & $34.08\pm0.78$ & \bestSc{$\mathbf{69.48}\pm\mathbf{1.57}$} \\
& 0.5 & $40.09\pm20.37$ & $42.05\pm10.52$ & $27.43\pm3.92$ & $29.72\pm16.02$ & $43.57\pm2.32$ & $26.22\pm1.95$ & \bestSc{$\mathbf{47.84}\pm\mathbf{1.52}$} & \secondbestSc{$44.36\pm1.07$} \\
& 2.0 & $8.91\pm2.28$ & $25.69\pm10.70$ & $11.88\pm9.38$ & $32.20\pm1.05$ & \secondbestSc{$44.32\pm4.58$} & $8.46\pm0.29$ & $6.97\pm0.04$ & \bestSc{$\mathbf{44.36}\pm\mathbf{2.35}$} \\
& 5.0 & $5.25\pm0.50$ & $5.42\pm0.29$ & $5.12\pm0.18$ & $5.44\pm0.08$ & \secondbestSc{$46.05\pm20.73$} & $4.31\pm0.04$ & $5.36\pm0.16$ & \bestSc{$\mathbf{54.01}\pm\mathbf{10.25}$} \\
\midrule

\multirow{4}{*}{\shortstack{Walker2d\\Friction}}
& 0.1 & $7.88\pm1.88$ & $5.72\pm0.23$ & \secondbestSc{$29.60\pm24.90$} & $5.65\pm0.06$ & $28.23\pm9.13$ & $9.32\pm0.22$ & $8.71\pm0.14$ & \bestSc{$\mathbf{61.66}\pm\mathbf{4.70}$} \\
& 0.5 & $63.94\pm20.40$ & $66.26\pm3.03$ & $45.01\pm18.98$ & $68.81\pm1.12$ & \bestSc{$\mathbf{76.96}\pm\mathbf{1.99}$} & $64.43\pm0.98$ & \secondbestSc{$70.36\pm0.76$} & $70.16\pm2.25$ \\
& 2.0 & $39.06\pm17.36$ & $65.40\pm7.13$ & $67.89\pm1.66$ & $72.91\pm0.37$ & \bestSc{$\mathbf{73.74}\pm\mathbf{0.49}$} & $48.38\pm3.14$ & $52.66\pm1.20$ & \secondbestSc{$73.05\pm0.02$} \\
& 5.0 & $10.07\pm4.91$ & $5.39\pm0.03$ & $5.76\pm0.84$ & $5.36\pm0.28$ & \secondbestSc{$27.38\pm3.87$} & $4.50\pm0.20$ & $6.19\pm0.41$ & \bestSc{$\mathbf{36.44}\pm\mathbf{8.23}$} \\
\midrule

\multirow{4}{*}{\shortstack{Hopper\\Gravity}}
& 0.1 & $27.82\pm13.41$ & $13.10\pm0.98$ & $15.59\pm6.09$ & $23.40\pm11.62$ & \bestSc{$\mathbf{36.25}\pm\mathbf{1.50}$} & $25.59\pm0.34$ & $29.37\pm0.41$ & \secondbestSc{$36.02\pm0.33$} \\
& 0.5 & $28.54\pm12.77$ & $16.24\pm7.89$ & $23.00\pm14.87$ & $12.86\pm0.18$ & $33.57\pm6.71$ & $34.92\pm1.27$ & \secondbestSc{$63.28\pm0.39$} & \bestSc{$\mathbf{68.25}\pm\mathbf{3.03}$} \\
& 2.0 & $11.84\pm2.37$ & $16.10\pm1.64$ & $18.62\pm6.88$ & $14.65\pm2.47$ & $23.79\pm2.09$ & $14.72\pm0.15$ & \bestSc{$\mathbf{31.38}\pm\mathbf{0.19}$} & \secondbestSc{$26.94\pm0.59$} \\
& 5.0 & $7.36\pm0.13$ & $8.12\pm0.16$ & \bestSc{$\mathbf{9.08}\pm\mathbf{1.15}$} & $7.90\pm1.27$ & $8.06\pm0.03$ & \secondbestSc{$9.00\pm0.04$} & $8.55\pm0.11$ & $8.80\pm0.02$ \\
\midrule

\multirow{4}{*}{\shortstack{Hopper\\Friction}}
& 0.1 & $25.55\pm2.69$ & $24.16\pm4.50$ & $18.64\pm3.37$ & $26.13\pm4.24$ & \secondbestSc{$51.19\pm2.56$} & $39.32\pm0.89$ & $36.65\pm1.56$ & \bestSc{$\mathbf{52.94}\pm\mathbf{7.71}$} \\
& 0.5 & $25.22\pm4.48$ & $23.56\pm1.68$ & $19.60\pm15.45$ & $26.94\pm2.86$ & \secondbestSc{$41.34\pm0.49$} & $39.82\pm1.46$ & $36.43\pm0.52$ & \bestSc{$\mathbf{41.97}\pm\mathbf{6.51}$} \\
& 2.0 & $10.32\pm0.06$ & $10.15\pm0.06$ & $9.89\pm0.20$ & $10.15\pm0.03$ & $11.00\pm0.14$ & \secondbestSc{$11.09\pm0.02$} & \bestSc{$\mathbf{14.06}\pm\mathbf{0.15}$} & $11.08\pm0.02$ \\
& 5.0 & $7.90\pm0.06$ & $7.93\pm0.01$ & $7.80\pm1.04$ & $7.86\pm0.05$ & $8.07\pm0.04$ & $7.94\pm0.01$ & \bestSc{$\mathbf{8.27}\pm\mathbf{0.00}$} & \secondbestSc{$8.24\pm0.04$} \\
\midrule
\multicolumn{2}{c|}{Total}
& $875.88$ & $901.52$ & $779.14$ & $890.62$
& \secondbestSc{$1427.26$} & $865.91$ & $908.92$
& \bestSc{$\mathbf{1504.16}$} \\
\bottomrule
\end{tabular}}
\end{table}

\subsection{Main Results}

We report the main results on MuJoCo tasks featuring gravity and friction shifts in \Cref{tab:mujoco_friction_gravity}. We evaluate two distinct applications of the proposed framework: \algname-Planner, which employs energy-guided diffusion directly for inference-time planning; and \algname-Policy, which utilizes the generated energy-guided trajectories as synthetic data for downstream policy optimization.

\Cref{tab:mujoco_friction_gravity} demonstrates that \algname achieves strong overall performance across both MuJoCo gravity and friction shifts. Among all tested methods, \algname-Policy obtains the highest aggregate normalized score, elevating the total score from $1427.26$ (achieved by the strongest baseline, MOBODY) to $1504.16$. This substantial improvement suggests that trajectory-level energy-guided generation provides significantly more valuable synthetic data than the transition-level model rollouts utilized by MOBODY under off-dynamics shifts. Furthermore, compared to standard offline RL baselines, \algname-Policy substantially outperforms IQL, improving the aggregate score from $901.52$ to $1504.16$. Since the downstream optimization of \algname-Policy is based on IQL, this performance gap confirms that the substantial gain is not simply attributable to the underlying offline RL backbone.

We also compare \algname{}-Planner with the vanilla Diffuser baseline to quantify the effectiveness of the energy guidance mechanism in diffusion-based trajectory generation. The vanilla Diffuser baseline trains a trajectory diffusion model on the combined source and target dataset without employing the proposed energy guidance. \algname{}-Planner improves the aggregate score over Diffuser from $865.91$ to $908.92$, demonstrating that energy guidance significantly enhances diffusion planning under dynamics shifts. Collectively, these findings highlight that diffusion modeling alone is insufficient in the off-dynamics setting, and the proposed energy guidance is critical for producing trajectories that are useful for both planning and subsequent policy learning.

\subsection{Ablation Study}

In this section, we present additional experiments to offer deeper insights of our proposed \algname.

\begin{table}[h]
\centering
\caption{Runtime analysis for Walker2d-Gravity-2.0 and HalfCheetah-Gravity-2.0.
Experiments are conducted on one NVIDIA RTX A5000 GPU with 24GB memory.
\textbf{Full} reports the whole cost of \algname. \textbf{Reuse} reports the extra cost when diffusion models are already available.}
\label{tab:runtime_adaptation}
\begin{tabular}{lcc|cc}
\toprule
\multirow{2}{*}{Method}
& \multicolumn{2}{c|}{Walker2d-Gravity-2.0}
& \multicolumn{2}{c}{HalfCheetah-Gravity-2.0} \\
\cmidrule(lr){2-3}\cmidrule(lr){4-5}
& Full & Reuse
& Full  & Reuse \\
\midrule
IQL
& $\sim 3.0$h & $\sim 3.0$h
& $\sim 3.0$h & $\sim 3.0$h \\

MOBODY
& $\sim 7.8$h & $\sim 7.8$h
& $\sim 6.1$h & $\sim 6.1$h \\

Diffuser
& $\sim 7.6$h & $\sim 7.6$h
& $\sim 5.5$h & $\sim 5.5$h \\

\algname{}-Planner
& $\sim 7.6$h & $\mathbf{\sim 3.3}$h
& $\sim 5.6$h & $\mathbf{\sim 1.9}$h \\

\algname{}-Policy
& $\sim 9.1$h & $\sim 4.8$h
& $\sim 7.4$h & $\sim 3.7$h \\
\bottomrule
\end{tabular}
\end{table}

\paragraph{Runtime analysis.}
We analyzed the computational cost of \algname. The primary efficiency advantage of \algname comes from the ability to reuse the source-domain trajectory diffusion model across different target shifts. As discussed in the experimental setup, the source diffusion model is trained once on the source dataset, while only the target-specific energy guidance functions need to be learned for each shifted target environment. \Cref{tab:runtime_adaptation} reports the runtime for two representative MuJoCo tasks. The \textbf{Full} column measures the cost incurred when all required components are trained for a new target shift. The \textbf{Reuse} column measures the additional cost required after the reusable diffusion model is already available. For baselines like IQL, Diffuser, and MOBODY, the runtime remains constant across tasks because they do not leverage component reuse and must retrain models for every new task. In contrast, \algname{} can reuse the source diffusion model and only retrain the target-specific energy module. Consequently, \algname-Planner reduces the adaptation cost from approximately $7.6$ hours to $3.3$ hours on Walker2d, and from approximately $5.6$ hours to $1.9$ hours on HalfCheetah. \algname-Policy also benefits from this reuse capability. These results strongly support the fast-adaptation design of \algname.

\begin{table}[h]
\centering
\caption{Effectiveness of guided synthetic data on HalfCheetah and Walker2d. All variants train IQL with the indicated data source; Diffuser and \algname{} use the same synthetic-data budget and are combined with the target dataset. Scores are mean $\pm$ std over three seeds.}
\label{tab:guided_synthetic_data}
\small
\setlength{\tabcolsep}{3pt}
\begin{tabular}{ccccccc}
\toprule
Shift Type & Level & Source & Target & Source+Target & Diffuser & \algname \\
\midrule
\multirow{4}{*}{\shortstack{HalfCheetah\\Gravity}}
& 0.1 & $8.87\pm1.24$ & $12.07\pm0.99$ & $9.62\pm4.27$ & $7.77\pm8.70$ & $\mathbf{20.04}\pm\mathbf{3.52}$ \\
& 0.5 & $\mathbf{45.01}\pm\mathbf{3.54}$ & $1.52\pm0.17$ & $44.23\pm2.93$ & $-0.19\pm0.29$ & $2.54\pm0.59$ \\
& 2.0 & $31.43\pm0.42$ & $27.77\pm6.53$ & $31.34\pm1.68$ & $13.41\pm7.26$ & $\mathbf{41.66}\pm\mathbf{5.84}$ \\
& 5.0 & $2.80\pm0.77$ & $73.81\pm0.12$ & $44.00\pm23.13$ & $55.10\pm15.36$ & $\mathbf{75.34}\pm\mathbf{1.63}$ \\
\midrule
\multirow{4}{*}{\shortstack{HalfCheetah\\Friction}}
& 0.1 & $10.59\pm0.27$ & $63.76\pm0.21$ & $26.39\pm11.35$ & $62.47\pm0.44$ & $\mathbf{64.31}\pm\mathbf{0.39}$ \\
& 0.5 & $\mathbf{76.15}\pm\mathbf{0.26}$ & $44.86\pm8.99$ & $69.80\pm0.64$ & $4.78\pm0.22$ & $52.06\pm6.41$ \\
& 2.0 & $\mathbf{52.02}\pm\mathbf{0.24}$ & $35.95\pm2.36$ & $46.04\pm2.04$ & $5.25\pm0.70$ & $49.15\pm4.03$ \\
& 5.0 & $16.52\pm3.55$ & $64.11\pm5.14$ & $44.96\pm6.78$ & $5.74\pm5.39$ & $\mathbf{68.56}\pm\mathbf{1.94}$ \\
\midrule
\multirow{4}{*}{\shortstack{Walker2d\\Gravity}}
& 0.1 & $16.59\pm1.25$ & $67.29\pm1.19$ & $16.04\pm7.60$ & $44.49\pm3.35$ & $\mathbf{69.48}\pm\mathbf{1.57}$ \\
& 0.5 & $16.18\pm4.29$ & $36.30\pm4.19$ & $42.05\pm10.52$ & $12.82\pm5.33$ & $\mathbf{44.36}\pm\mathbf{1.07}$ \\
& 2.0 & $8.68\pm0.68$ & $41.17\pm3.46$ & $25.69\pm10.70$ & $19.18\pm4.46$ & $\mathbf{44.36}\pm\mathbf{2.35}$ \\
& 5.0 & $2.22\pm1.64$ & $\mathbf{55.78}\pm\mathbf{14.38}$ & $5.42\pm0.29$ & $5.27\pm0.18$ & $54.01\pm10.25$ \\
\midrule
\multirow{4}{*}{\shortstack{Walker2d\\Friction}}
& 0.1 & $5.39\pm0.05$ & $52.97\pm9.10$ & $5.72\pm0.23$ & $10.27\pm1.90$ & $\mathbf{61.66}\pm\mathbf{4.70}$ \\
& 0.5 & $66.78\pm0.59$ & $62.21\pm6.37$ & $66.26\pm3.03$ & $25.25\pm8.28$ & $\mathbf{70.16}\pm\mathbf{2.25}$ \\
& 2.0 & $45.66\pm5.00$ & $72.96\pm0.61$ & $65.40\pm7.13$ & $17.71\pm2.88$ & $\mathbf{73.05}\pm\mathbf{0.02}$ \\
& 5.0 & $4.82\pm0.06$ & $22.77\pm3.07$ & $5.39\pm0.03$ & $6.56\pm0.64$ & $\mathbf{36.44}\pm\mathbf{8.23}$ \\
\bottomrule
\end{tabular}
\end{table}

\begin{wrapfigure}{r}{0.45\textwidth}
    \centering
    \vspace{-10pt}
    \includegraphics[width=\linewidth]{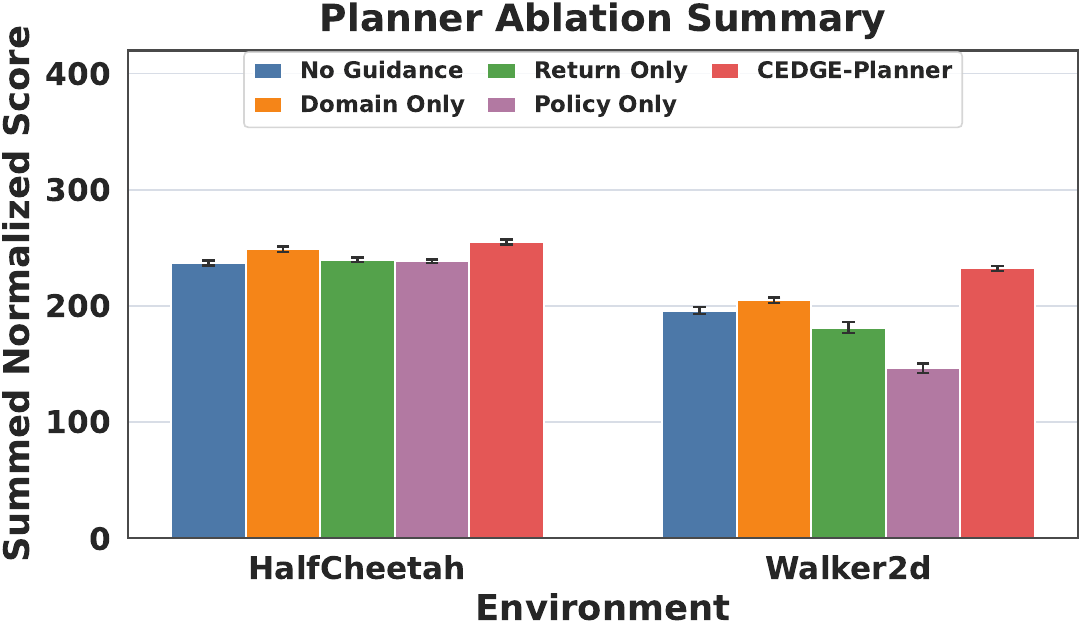}
    \caption{
    Planner-only ablation of energy guidance on HalfCheetah and Walker2d.
    }
    \label{fig:guided_synthetic_data_robot_sum}
    \vspace{-10pt}
\end{wrapfigure}

\paragraph{Contribution of energy guidance in planning.}
We then study the role of different energy guidance terms in the planning setting. This ablation is conducted on \algname{}-Planner only. At each target environment step, all variants use the same source trajectory diffusion model to sample trajectory candidates conditioned on the current state. They also use the same planning protocol and the same return energy score to select the trajectory for action execution. The only difference is which guidance term is used during reverse diffusion sampling. We compare the full \algname{}-Planner with four variants. \textbf{No Guidance} removes all energy guidance terms. \textbf{Domain Only},\textbf{ Return Only}, and \textbf{Policy Only} use only the domain energy, return energy, and policy energy respectively.
As shown in \Cref{fig:guided_synthetic_data_robot_sum}, the full \algname{}-Planner achieves the best performance across HalfCheetah and Walker2d shifts in this ablation. Adding a single energy term can improve performance in some settings, but it does not provide consistent gains across tasks and shift levels. This suggests that no individual energy term is sufficient by itself. The full planner benefits from the synergy of these energy terms. Detailed results can be found in \Cref{tab:energy_components} in \Cref{app:only_guidance}.

\paragraph{Effectiveness of guided synthetic data in policy optimization.}
We then evaluate whether energy-guided trajectories provide useful synthetic data for policy optimization. We fix IQL as the offline RL policy for all variants and only change the data used to train IQL. \textbf{Source} uses only the source domain D4RL dataset, \textbf{Target} uses only the limited target dataset, and \textbf{Source+Target} uses their combination. \textbf{Diffuser} uses synthetic trajectories generated by the vanilla Diffuser and combined with the target dataset. As shown in \Cref{tab:guided_synthetic_data}, \algname{} outperforms Diffuser in all 16 settings. This shows that vanilla diffusion generation alone is not enough under dynamics shift and energy guidance is important for generating useful synthetic trajectories. \algname{} also outperforms Target in 15 out of 16 settings, showing that the generated trajectories provide useful information beyond the scarce target dataset. Compared with Source+Target, \algname{} performs better in 14 out of 16 settings, indicating that energy-guided synthetic data is usually more effective than directly combining the source and target datasets. Overall, these results show that the policy improvement comes from the quality of the energy-guided synthetic trajectories.

\begin{figure}[h]
    \centering
    \begin{subfigure}{0.45\linewidth}
        \centering
        \includegraphics[width=\linewidth]{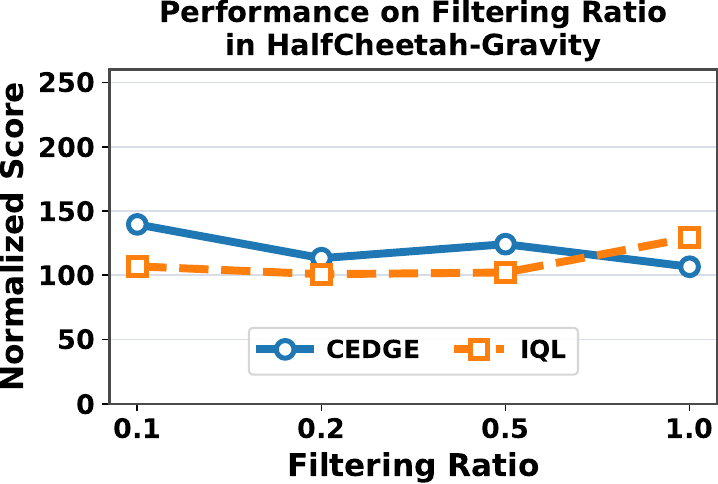}
        \caption{Gravity shifts}
        \label{fig:filter_control_gravity}
    \end{subfigure}
    \hfill
    \begin{subfigure}{0.45\linewidth}
        \centering
        \includegraphics[width=\linewidth]{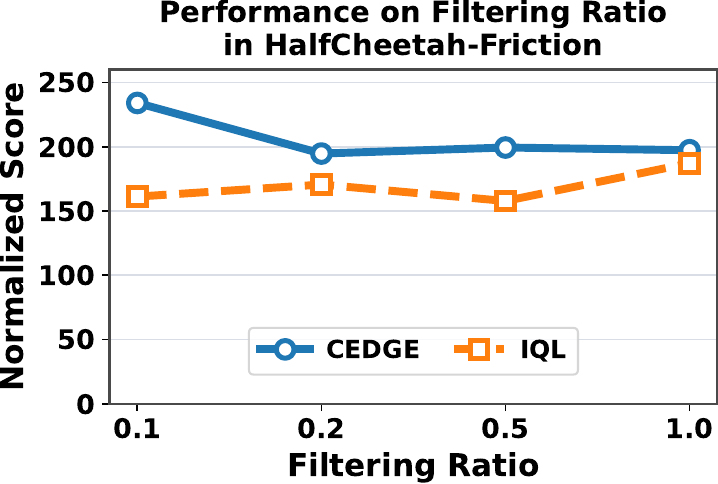}
        \caption{Friction shifts}
        \label{fig:filter_control_friction}
    \end{subfigure}
    \caption{Filtering ratio performance on HalfCheetah.
    For each shift type, we report the sum of normalized scores over four shift levels $\{0.1,0.5,2.0,5.0\}$ under filtering ratios $\{0.1,0.2,0.5,1.0\}$.
    Both synthetic and source data are filtered with the same criterion and used to train the IQL policy combined with the limited target dataset.}
    \label{fig:filter_control}
\end{figure}

\paragraph{Is the gain merely due to data filtering in policy optimization?}
We further check whether the improvement of \algname{}-Policy comes only from filtering. We apply the same filtering criterion to both the generated synthetic dataset and the original source dataset. For each data source, we retain the top $\{0.1,0.2,0.5,1.0\}$
fraction and train the same downstream IQL policy together with the limited target dataset. \Cref{fig:filter_control} reports the normalized scores over four shift levels for HalfCheetah gravity and friction shifts. If filtering alone explains the improvement, then filtered source data should achieve similar aggregate performance to filtered synthetic data under the same filtering ratio. In the friction shifts, filtered synthetic data consistently outperforms filtered source data across all filtering ratios. In the gravity shifts, filtered synthetic data also performs better under most filtering ratios. These results show that filtering is useful, but it cannot fully explain the improvement of \algname-Policy. The gain mainly comes from generating synthetic trajectories rather than simply filtering the original source dataset. More detailed results can be found in \Cref{tab:filter_control} in \Cref{app:filter_ratio}.

\section{Conclusion and Limitations}
In this paper, we proposed \algname, an energy-guided diffusion generation framework for offline off-dynamics reinforcement learning. \algname learns a trajectory-level diffusion from abundant source-domain data and generates energy-guided trajectories. 
We derive the guidance by reweighting the source trajectory distribution to reduce its mismatch with the desired target-domain trajectory distribution, which yields three decomposed energy terms including domain energy, return energy, and policy energy. \algname supports two downstream uses of the generated energy-guided trajectories. First, the guided diffusion model can be used directly as a planner. Second, the generated trajectories can serve as synthetic offline data for policy optimization. Experiments on the ODRL benchmark under gravity and friction dynamics shifts show improved performance compared with baselines. Our work has several limitations. Its performance depends on learned guidance modules, which may be inaccurate with very limited target data. Experiments are limited to simulated MuJoCo tasks, and the method adds computation for energy-module training and diffusion sampling.

\clearpage

\appendix

\section{Algorithm Details}
\label{app:technical_details}

This section provides the technical details of the source-domain trajectory diffusion model, the energy guidance functions, and the energy-guided reverse diffusion generation. We first train a trajectory diffusion model on source-domain trajectories, which provides a reusable generator for the source trajectory distribution. For each new target environment, we keep the diffusion model fixed and train target-specific energy modules using the available source and target offline datasets. During reverse diffusion sampling, the energy functions are evaluated on the current noisy trajectory sample, and their gradients are added to the diffusion score to guide the generation process. We summarize the process in \Cref{alg:diffusion_training}, \Cref{alg:energy_training} and \Cref{alg:energy_guided_sampling}.

\begin{algorithm}[htbp]
\caption{Training the Source-Domain Trajectory Diffusion Model}
\label{alg:diffusion_training}
\begin{algorithmic}[1]
\Require Source dataset $\mathcal D_{\mathrm{src}}$, denoiser $D_\theta$, diffusion schedule $\{\beta_k\}_{k=1}^{T}$.
\For{\textbf{epoch} $=1,\dots,E_{\mathrm{diff}}$}
    \State Sample a source trajectory batch $\tau \sim \mathcal D_{\mathrm{src}}$
    \State Sample diffusion step $k \sim \mathrm{Uniform}(\{1,\dots,T\})$ and noise $\epsilon \sim \mathcal N(0,I)$
    \State Construct the noisy trajectory $\tau_k$ according to the forward diffusion process
    \State Update the denoiser $D_\theta$ using the diffusion loss in \eqref{eq:diffusion_loss}
\EndFor
\end{algorithmic}
\end{algorithm}

\begin{algorithm}[htbp]
\caption{Training Energy Guidance Functions for a Target Environment}
\label{alg:energy_training}
\begin{algorithmic}[1]
\Require Source dataset $\mathcal D_{\mathrm{src}}$, target dataset $\mathcal D_{\mathrm{trg}}$, fixed denoiser $D_\theta$, domain classifiers $p_{\theta_{\mathrm{SAS}}}, p_{\theta_{\mathrm{SA}}}$, return predictor $J_\psi$, Gaussian behavior policy $\pi_\omega$.
\For{\textbf{epoch} $=1,\dots,E_{\mathrm{energy}}$}
    \State Sample source and target transition batches from $\mathcal D_{\mathrm{src}}$ and $\mathcal D_{\mathrm{trg}}$
    \State Update $p_{\theta_{\mathrm{SAS}}}$ and $p_{\theta_{\mathrm{SA}}}$ using \eqref{eq:loss_sas} and \eqref{eq:loss_sa}.
    \State Sample trajectory batches from $\mathcal D_{\mathrm{src}} \cup \mathcal D_{\mathrm{trg}}$
    \State Update $J_\psi$ using \eqref{eq:return_energy}
    \State Sample source state-action pairs from $\mathcal D_{\mathrm{src}}$
    \State Update $\pi_\omega$ using \eqref{eq:policy_energy}
\EndFor
\end{algorithmic}
\end{algorithm}

\begin{algorithm}[htbp]
\caption{Energy-Guided Reverse Diffusion Sampling}
\label{alg:energy_guided_sampling}
\begin{algorithmic}[1]
\Require Fixed denoiser $D_\theta$, guidance functions $\mathcal E_1,\mathcal E_2,\mathcal E_3$, guidance weights $\lambda_1,\lambda_2,\lambda_3$, guidance scales $\{\rho_k\}_{k=1}^{T}$, diffusion schedule $\{\beta_k\}_{k=1}^{T}$, number of samples $N$.
\State Let $\alpha_k=1-\beta_k$ and $\bar{\alpha}_k=\prod_{i=1}^{k}\alpha_i$
\State Sample initial noisy trajectories $\{\tau_T^{(n)}\}_{n=1}^{N}\sim \mathcal N(0,I)$
\For{$k=T,\dots,1$}
    \State Predict the clean trajectory estimate:
    $
    \hat{\tau}_0^{(n)} = D_\theta(\tau_k^{(n)},k)$, where $n=1,\dots,N.
    $
    \State Convert denoiser into the diffusion score:
    $
    s_\theta(\tau_k^{(n)},k)
    =-\frac{\tau_k^{(n)}-\sqrt{\bar{\alpha}_k}\hat{\tau}_0^{(n)}}{1-\bar{\alpha}_k}.
    $
    \State Compute the weighted guidance energy:
    $$
    \mathcal E(\tau_k^{(n)},k)
    =\lambda_1\mathcal E_1(\tau_k^{(n)},k)+
    \lambda_2\mathcal E_2(\tau_k^{(n)},k)+\lambda_3\mathcal E_3(\tau_k^{(n)},k).
    $$
    \State Compute the energy guidance score:
    $
    h_k^{(n)}=-\nabla_{\tau_k^{(n)}}\mathcal E(\tau_k^{(n)},k).
    $
    \State Add the energy guidance score to the diffusion score:
    $
    \tilde{s}_\theta(\tau_k^{(n)},k)
    =s_\theta(\tau_k^{(n)},k)
    +\rho_k h_k^{(n)}.
    $
    \State Compute the guided reverse mean:
    $
    \tilde{\mu}_k^{(n)}
    =\frac{1}{\sqrt{\alpha_k}}
    \left(
    \tau_k^{(n)}+\beta_k \tilde{s}_\theta(\tau_k^{(n)},k)
    \right).
    $
    \State Sample $\tau_{k-1}^{(n)} \sim \mathcal N(\tilde{\mu}_k^{(n)}, \sigma_k^2 I)$
\EndFor
\end{algorithmic}
\end{algorithm}

\section{Experimental Details}
\label{app:experimental_details}

\subsection{Environments and datasets}
\label{app:env_details}

We evaluate on the ODRL benchmark \citep{lyu2024odrlbenchmarkoffdynamicsreinforcement}.  Following the ODRL benchmark protocol, the source domain is the original MuJoCo environment, while each target domain is constructed by modifying the simulator dynamics parameters. The source and target domains share the same state space, action space, reward function, and initial-state distribution, and differ only in the transition dynamics. For MuJoCo locomotion tasks, we consider four environments: HalfCheetah, Ant, Walker2d, and Hopper.

We evaluate two representative types of dynamics shifts in MuJoCo from ODRL: gravity shifts and friction shifts. For gravity shifts, we modify the magnitude of the gravity vector while keeping its direction unchanged. 
For friction shifts, we modify all MuJoCo friction components associated with the robot-environment contact, including static, dynamic, and rolling friction. In both cases, the target environment is obtained by multiplying the corresponding source-domain simulator parameter by a factor in $\{0.1,0.5,2.0,5.0\}$. These factors cover both mild and severe dynamics mismatch and follow the task construction used in ODRL.

We also evaluate our proposed \algname{} under kinematic and morphology shifts in MuJoCo tasks. For kinematic shifts, ODRL modifies the rotation range of selected joints, which simulates broken or constrained joints and changes the feasible motion patterns of the robot. For morphology shifts, ODRL modifies the size of selected limbs or torso parts while keeping the state and action spaces unchanged. Following the ODRL protocol, we evaluate \algname{} on the medium and hard levels of these shifts across HalfCheetah, Ant, Walker2d, and Hopper.

For the offline datasets, we use the original D4RL medium-level datasets collected from the unmodified MuJoCo environments as the source datasets. Each source dataset contains $1$M transitions. For each shifted target environment, we use the limited target-domain dataset provided by ODRL, which contains $5{,}000$ transitions collected from the corresponding shifted environment by a medium-performance SAC policy \citep{haarnoja2018soft}. All methods are trained only on the provided offline source and target datasets and are evaluated in the shifted target environments.

The normalized score, similar to D4RL \citep{fu2020d4rl}, is computed as
\begin{align}
\mathrm{NS}=\frac{J-J_{\mathrm{random}}}{J_{\mathrm{expert}}-J_{\mathrm{random}}}\times 100,
\end{align}
where $J$ is the evaluated return in the target domain, and $J_{\mathrm{random}}$ and $J_{\mathrm{expert}}$ are the random-policy and expert-policy returns in the same target environment.

\subsection{ODRL task configuration}

ODRL constructs target-domain locomotion environments by editing the MuJoCo XML files.

\paragraph{Gravity Shift.}
Following the ODRL benchmark \citep{lyu2024odrlbenchmarkoffdynamicsreinforcement}, gravity shifts are generated by editing the \texttt{gravity} attribute in the MuJoCo \texttt{option} element.
Only the magnitude of the gravity vector is modified, while its direction is kept unchanged.
For example, to construct \texttt{halfcheetah-gravity-0.5}, the gravity in the target domain is set to $0.5$ times the gravity in the source domain:
\begin{tcolorbox}[
  colback=black!2,
  colframe=black!30,
  boxrule=0.4pt,
  left=6pt,
  right=6pt,
  top=4pt,
  bottom=4pt
]
\begin{lstlisting}[style=xmlstyle]
# gravity
<option gravity="0 0 -4.905" timestep="0.01"/>
\end{lstlisting}
\end{tcolorbox}

\paragraph{Friction Shift.}
Friction shifts are generated by modifying the \texttt{friction} attribute in the MuJoCo \texttt{geom} elements.
ODRL adjusts the frictional components to $\{0.1,0.5,2.0,5.0\}$ times the corresponding source-domain values.
The following example from ODRL illustrates a $5.0\times$ friction shift for Hopper:
\begin{tcolorbox}[
  colback=black!2,
  colframe=black!30,
  boxrule=0.4pt,
  left=6pt,
  right=6pt,
  top=4pt,
  bottom=4pt
]
\begin{lstlisting}[style=xmlstyle]
# torso
<geom friction="4.5" fromto="0 0 1.45 0 0 1.05" 
      name="torso_geom" size="0.05" type="capsule"/>

# thigh
<geom friction="4.5" fromto="0 0 1.05 0 0 0.6" 
      name="thigh_geom" size="0.05" type="capsule"/>

# leg
<geom friction="4.5" fromto="0 0 0.6 0 0 0.1" 
      name="leg_geom" size="0.04" type="capsule"/>

# foot
<geom friction="10.0" fromto="-0.13 0 0.1 0.26 0 0.1" 
      name="foot_geom" size="0.06" type="capsule"/>
\end{lstlisting}
\end{tcolorbox}

\paragraph{Kinematic Shift.}
Kinematic shifts are generated by modifying the rotation ranges of selected joints in the MuJoCo XML files. Following ODRL \citep{lyu2024odrlbenchmarkoffdynamicsreinforcement}, these shifts simulate broken or constrained joints, so that some motion patterns become infeasible in the target domain. Each kinematic shift has three difficulty levels: easy, medium, and hard. The harder levels impose narrower rotation ranges and therefore induce stronger transition-dynamics mismatch.

\paragraph{Morphology Shift.}
Morphology shifts are generated by modifying the size or geometry of selected robot body parts in the MuJoCo XML files. Following ODRL \citep{lyu2024odrlbenchmarkoffdynamicsreinforcement}, each robot has two types of morphology changes with three difficulty levels: easy, medium, and hard.
For example, Ant includes morphology shifts on the front two legs or all legs; HalfCheetah includes shifts on the thigh or torso; Hopper includes shifts on the foot or torso; and Walker2d includes shifts on the leg or torso.

\subsection{Baselines}
\label{app:baseline_details}

We compare our method with standard offline RL baselines and off-dynamics offline RL baselines. For standard offline RL baselines, we include IQL \citep{kostrikov2021offline} and TD3-BC \citep{fujimoto2021minimalist}.  IQL learns an expectile value function and extracts the policy via advantage-weighted behavioral cloning, while TD3-BC adds a behavior cloning regularizer to the policy update.  For off-dynamics offline RL baselines, we include DARA \citep{liu2022dara}, BOSA \citep{liu2024beyond}, and MOBODY \citep{guo2026mobody}. DARA uses domain classifiers to estimate the source-target dynamics discrepancy and performs dynamics-aware reward augmentation. BOSA addresses out-of-distribution state-action pairs and dynamics mismatch through support-constrained policy and value optimization. MOBODY is a model-based off-dynamics offline RL method that learns a target-aware dynamics model and generates transition-level rollouts for policy improvement. We also include a Diffuser baseline \citep{janner2022planning}, which trains a trajectory diffusion model on the combined source and target offline datasets without the proposed energy guidance.

\paragraph{IQL.}
Implicit Q-Learning (IQL) \citep{kostrikov2021offline} is an offline RL algorithm that learns a value function, a Q-function, and then extracts a policy by weighted behavior cloning. Let $\tau_{\mathrm{exp}}$ be the expectile
parameter and define $\ell_{\tau_{\mathrm{exp}}}(u)=|\tau_{\mathrm{exp}}-\mathbf{1}\{u<0\}|u^2$. IQL trains the value function and Q-function with
\begin{align}
\mathcal L_V(\psi)
&=\mathbb E_{(s,a)\sim\mathcal D}
\left[
\ell_{\tau_{\mathrm{exp}}}
\left(Q_\theta(s,a)-V_\psi(s)\right)
\right], \\
\mathcal L_Q(\theta)
&=\mathbb E_{(s,a,r,s')\sim\mathcal D}
\left[
\left(r+\gamma V_{\bar{\psi}}(s')-Q_\theta(s,a)\right)^2
\right],
\end{align}
where $\mathcal D$ is the training dataset. The policy is trained by advantage-weighted behavior cloning:
\begin{align}
\mathcal L_\pi(\phi)
=-\mathbb E_{(s,a)\sim\mathcal D}
\left[
\exp\left(\beta\left(Q_\theta(s,a)-V_\psi(s)\right)\right)
\log \pi_\phi(a\mid s)
\right].
\end{align}
In the main comparison, IQL is trained on
$\mathcal D_{\mathrm{src}}\cup\mathcal D_{\mathrm{trg}}$.

\paragraph{TD3-BC.}
TD3-BC \citep{fujimoto2021minimalist} is an offline variant of TD3 with an additional behavior cloning term in the actor loss. The critic target is
\begin{align}
y=r+\gamma
\min_{i=1,2}
Q_{\bar{\theta}_i}
\left(s', \pi_{\bar{\phi}}(s')+\epsilon\right),
\end{align}
where $\epsilon$ is the target-policy smoothing noise. The actor is trained with
\begin{align}
\mathcal L_\pi(\phi)
= -\lambda
\mathbb E_{s\sim\mathcal D}
\left[
Q_{\theta_1}(s,\pi_\phi(s))
\right]
+
\mathbb E_{(s,a)\sim\mathcal D}
\left[
\|\pi_\phi(s)-a\|_2^2
\right],
\end{align}
where $\lambda=\alpha/\mathbb E_{s\sim\mathcal D}[|Q_{\theta_1}(s,\pi_\phi(s))|]$ normalizes the Q term, and $\alpha$ controls the relative strength of Q maximization and behavior cloning.
In the main comparison, TD3-BC is trained on
$\mathcal D_{\mathrm{src}}\cup\mathcal D_{\mathrm{trg}}$.

\paragraph{DARA.}
DARA \citep{liu2022dara} uses domain classifiers to estimate the source-target dynamics discrepancy of source transitions under dynamics shift.
Following the classifier-ratio form used in off-dynamics RL \citep{eysenbach2020off}, DARA trains a transition classifier, $p_{\theta_{\mathrm{SAS}}}(d\mid s,a,s')$, and a state-action classifier, $p_{\theta_{\mathrm{SA}}}(d\mid s,a)$, where $d\in\{\mathrm{src},\mathrm{trg}\}$.
The source-over-target dynamics log-ratio is estimated as
\begin{align}
\widehat{\Delta}_{\mathrm{dyn}}(s,a,s')
=\log
\frac{
p_{\theta_{\mathrm{SAS}}}(\mathrm{src}\mid s,a,s')
}{
p_{\theta_{\mathrm{SAS}}}(\mathrm{trg}\mid s,a,s')
}
-\log
\frac{
p_{\theta_{\mathrm{SA}}}(\mathrm{src}\mid s,a)
}{
p_{\theta_{\mathrm{SA}}}(\mathrm{trg}\mid s,a)
}.
\end{align}
DARA then modifies the source-domain rewards by
\begin{align}
\tilde r(s,a,s')
=r(s,a,s')-\eta
\widehat{\Delta}_{\mathrm{dyn}}(s,a,s'),
\end{align}
where $\eta>0$ controls the strength of the source-likeness penalty. This lowers the reward of source transitions that are estimated to be more source-like than target-like. The policy is then trained on the reward-augmented source data together with the limited target data.

\paragraph{BOSA.}
BOSA \citep{liu2024beyond} addresses the off-dynamics offline RL problem from two aspects. First, it handles out-of-distribution (OOD) state-action pairs through supported policy optimization. Second, it addresses the OOD dynamics through supported value optimization with data filtering. Specifically, the policy is updated by maximizing
\begin{align}
\mathcal L_{\mathrm{actor}}
= \mathbb E_{s \sim \mathcal D_{\mathrm{src}} \cup \mathcal D_{\mathrm{trg}},\, a \sim \pi_\phi(\cdot \mid s)}
\left[
Q(s,a)
\right],
\quad
\mathrm{s.t.} \mathbb E_{s \sim \mathcal D_{\mathrm{src}} \cup \mathcal D_{\mathrm{trg}}}
\left[
\hat{\pi}_{\mathrm{offline}}(\pi_\phi(s)\mid s)
\right] > \epsilon,
\end{align}
where $\epsilon$ is a threshold and $\hat{\pi}_{\mathrm{offline}}$ is the learned behavior policy on the combined offline dataset. The value critic is updated with
\begin{align}
\mathcal L_{\mathrm{critic}}
&= \mathbb E_{(s,a)\sim \mathcal D_{\mathrm{src}}}
\left[
Q(s,a)
\right] \nonumber \\
&\quad + \mathbb E_{(s,a,r,s') \sim \mathcal D_{\mathrm{src}} \cup \mathcal D_{\mathrm{trg}},\,
a' \sim \pi_\phi(\cdot \mid s')}
\left[ I\left(\hat p_{\mathrm{trg}}(s' \mid s,a) > \epsilon' \right) \left(Q_{\theta_i}(s,a)-y\right)^2
\right],
\end{align}
where $I(\cdot)$ is the indicator function, $\hat p_{\mathrm{trg}}(s' \mid s,a)$ denotes the estimated target-domain dynamics model, and $\epsilon'$ is a threshold.

\paragraph{MOBODY.}
MOBODY \citep{guo2026mobody} is a model-based off-dynamics offline RL method that learns target-domain dynamics and uses model rollouts for policy optimization. It is motivated by the limitation that reward-regularization and data-filtering methods mainly rely on low-shift source transitions and may fail to explore high-return target-domain regions under large dynamics mismatch. To learn target dynamics from abundant source data and limited target data, MOBODY uses separate source and target action encoders, $\psi_{\mathrm{src}}$ and $\psi_{\mathrm{trg}}$, together with a shared state encoder $\phi_E$ and a shared transition model $\phi_T$:
\begin{align}
\hat{s}'_{\mathrm{src}}
&= \phi_T\left(\phi_E(s)+\psi_{\mathrm{src}}(\phi_E(s),a)\right), \\
\hat{s}'_{\mathrm{trg}}
&= \phi_T\left(\phi_E(s)+\psi_{\mathrm{trg}}(\phi_E(s),a)\right).
\end{align}
After learning the target dynamics, MOBODY generates synthetic rollout data and trains the policy on an enhanced dataset composed of DARA-regularized source data, limited target data, and generated rollout data. It further uses a target-Q-weighted behavior cloning loss to regularize the policy toward actions with high target-domain Q-values.

\paragraph{Diffuser.}
Diffuser \citep{janner2022planning} is a trajectory-level diffusion planning method for offline RL. It models state-action trajectories directly and generates plans through iterative denoising. In our experiments, the Diffuser baseline trains a trajectory denoiser on the combined source and target trajectory dataset $\mathcal D_{\mathrm{src}}\cup\mathcal D_{\mathrm{trg}}$ with the data-prediction diffusion objective
\begin{align}
\mathcal L_{\mathrm{diff}}(\theta)
= \mathbb E_{\tau_0\sim
\mathcal D_{\mathrm{src}}\cup\mathcal D_{\mathrm{trg}},\,
k,\epsilon}
\left[
\left\|
D_\theta(\tau_k,k)-\tau_0
\right\|_2^2
\right],
\end{align}
where
\begin{align}
\tau_k=\sqrt{\bar\alpha_k}\tau_0+\sqrt{1-\bar\alpha_k}\epsilon,
\quad
\epsilon\sim\mathcal N(0,I).
\end{align}
At evaluation time, Diffuser samples candidate trajectories by reverse denoising. The first state of each trajectory is constrained to the current state, and the first action of the selected trajectory is executed.

\subsection{Implementation Details}
\label{app:detailed_implement}
We implement \algname as an energy-guided trajectory diffusion model over normalized state-action sequences. The source domain trajectory diffusion model is parameterized by a temporal 1D U-Net denoiser and is trained only on source trajectories. Energy guidance is introduced through three learned energy modules, corresponding to domain energy, return energy, and policy energy. We use a discrete diffusion model with a cosine noise schedule, DDPM sampling, and $20$ reverse sampling steps. The denoiser and energy modules are trained with a batch size of $64$. For energy guided generation, the return and domain guidance weights are chosen from $\{0.1,0.5,1.0,2.0\}$, while the policy guidance weight is set to be $0.1$. Additional architecture and hyperparameter details are provided in \Cref{app:implement_details}.

\section{Hyperparameters and Architectures}
\label{app:implement_details}

This section summarizes the architectures and hyperparameters used in \algname{}. All trajectory-generation components operate on trajectory tensors containing states and actions. In the implementation, each trajectory segment is stored as a length-$H$ state-action sequence,
\begin{align*}
    \tau = (s_0,a_0,s_1,a_1,\ldots,s_{H-1},a_{H-1}).
\end{align*}
Trajectory segments are extracted from individual D4RL episodes, where episode boundaries are determined by terminal or timeout flags, and no segment crosses an episode boundary. Observations are normalized using a Gaussian normalizer fitted on the offline training data, while actions are kept in their original environment scale. During planning, generated actions are clipped to the valid action range $[-1, 1]$ before environment execution.

\paragraph{Diffusion model.}
The source-domain trajectory diffusion model is parameterized by a temporal 1D U-Net denoiser. The denoiser takes a noisy trajectory segment $\tau_k$ and diffusion timestep $k$ as input and predicts the clean trajectory segment $\tau_0$. In implementation, each trajectory segment is represented as a length-$H$ state-action tensor with shape $H \times (d_s+d_a)$, where $d_s$ and $d_a$ denote the state and action dimensions respectively. The model architecture is summarized in \Cref{tab:diffusion_architecture}. Task-specific sequence lengths and channel multipliers are reported in \Cref{tab:diffusion_task_configs}, and the diffusion training hyperparameters are listed in \Cref{tab:diffusion_hyperparameters}. The source-domain diffusion model is trained once and reused across target dynamics shifts.

\begin{table}[t]
\centering
\small
\setlength{\tabcolsep}{6pt}
\caption{Architecture of the trajectory diffusion denoiser.}
\label{tab:diffusion_architecture}
\begin{tabular}{lc}
\toprule
Component & Setting \\
\midrule
Denoiser & Temporal 1D U-Net \\
Input / output shape & $H \times (d_s+d_a)$ \\
Prediction target &  $\tau_0$ \\
Base channels & $32$ \\
Timestep embedding dimension & $32$ \\
Convolution kernel size & $5$ \\
\bottomrule
\end{tabular}
\end{table}

\begin{table}[t]
\centering
\small
\setlength{\tabcolsep}{8pt}
\caption{Task-specific diffusion horizon and channel multipliers.}
\label{tab:diffusion_task_configs}
\begin{tabular}{lcc}
\toprule
Environment & Horizon $H$ & Channel multipliers \\
\midrule
HalfCheetah & $4$ & $[1, 4, 2]$ \\
Ant & $4$ & $[1, 4, 2]$ \\
Walker2d & $32$ & $[1, 2, 2, 2]$ \\
Hopper & $32$ & $[1, 2, 2, 2]$ \\
\bottomrule
\end{tabular}
\end{table}

\begin{table}[t]
\centering
\small
\setlength{\tabcolsep}{8pt}
\caption{Training hyperparameters of the source-domain trajectory diffusion model.}
\label{tab:diffusion_hyperparameters}
\begin{tabular}{lc}
\toprule
Hyperparameter & Value \\
\midrule
Noise schedule & Cosine \\
Diffusion steps & $20$ \\
Optimizer & AdamW \\
Learning rate & $2\times 10^{-4}$ \\
Weight decay & $10^{-5}$ \\
AdamW betas & $(0.9,0.999)$ \\
Batch size & $64$ \\
Training steps & $5\times10^5$ \\
EMA decay & $0.9999$ \\
Learning-rate schedule & Cosine annealing \\
\bottomrule
\end{tabular}
\end{table}

\paragraph{Energy modules.}
For each target environment, we keep the source-domain trajectory diffusion model fixed and train target-specific energy modules using the available source and target offline datasets. The sampling energy is
\begin{align}
\mathcal E(\tau)
=
\lambda_1\mathcal E_1(\tau)
+
\lambda_2\mathcal E_2(\tau)
+
\lambda_3\mathcal E_3(\tau),
\end{align}
where $\mathcal E_1$, $\mathcal E_2$, and $\mathcal E_3$ denote the domain energy, return energy, and source behavior-policy energy, respectively.
The architectures of these energy modules are shown in
\Cref{tab:energy_architecture}, and their training and guidance hyperparameters are reported in \Cref{tab:energy_hyperparameters}. During reverse diffusion sampling, the energy functions are evaluated on the current noisy trajectory sample in the normalized trajectory space.

\begin{table}[t]
\centering
\small
\setlength{\tabcolsep}{5pt}
\caption{Architectures of the energy modules and reward model.}
\label{tab:energy_architecture}
\begin{tabular}{lll}
\toprule
Module & Input & Architecture \\
\midrule
Domain energy $\mathcal E_1$ &
$(s,a,s')$ and $(s,a)$ &
Two domain classifiers, two-layer MLPs, hidden size $256$ \\
Return energy $\mathcal E_2$ &
Trajectory $\tau$ &
Half U-Net, kernel size $3$ \\
Source behavior-policy energy $\mathcal E_3$ &
$(s,a)$ &
Gaussian behavior policy, hidden size $256$ \\
Reward annotator $\hat r_\eta$ &
$(s,a,s')$ &
Two-layer MLP, hidden size $256$ \\
\bottomrule
\end{tabular}
\end{table}

\begin{table}[t]
\centering
\small
\setlength{\tabcolsep}{7pt}
\caption{Training and guidance hyperparameters of the energy modules.}
\label{tab:energy_hyperparameters}
\begin{tabular}{lc}
\toprule
Hyperparameter & Value \\
\midrule
Energy optimizer & Adam \\
Energy learning rate & $2\times 10^{-4}$ \\
Energy weight decay & $10^{-4}$ \\
Energy training steps & $2\times 10^5$ \\
Return-energy batch size & $64$ \\
Domain-energy batch size & $64$ source + $64$ target \\
Domain source-target ratio & $1{:}1$ \\
Source behavior-policy batch size & $64$ source state-action pairs \\
Reward annotator optimizer & Adam \\
Reward annotator learning rate & $3\times 10^{-4}$ \\
Reward annotator training steps & $2\times 10^5$ \\
Reward annotator batch size & $64$ \\
Domain guidance weight $\lambda_1$ & $\{0.1, 0.5, 1.0, 2.0\}$ \\
Return guidance weight $\lambda_2$ & $\{0.1, 0.5, 1.0, 2.0\}$ \\
Source behavior-policy guidance weight $\lambda_3$ & $0.1$ \\
\bottomrule
\end{tabular}
\end{table}

\paragraph{\algname{}-Planner.}
\algname{}-Planner samples candidate trajectories conditioned on the current state and executes the first action of the selected trajectory.
Candidate trajectories are ranked by return energy.
Since the return energy is defined as the negative predicted return, selecting the trajectory with the lowest return energy corresponds to selecting the trajectory with the highest predicted return.
The planning hyperparameters are summarized in \Cref{tab:planning_hyperparameters}.

\begin{table}[t]
\centering
\small
\setlength{\tabcolsep}{7pt}
\caption{Planning hyperparameters for \algname{}-Planner.}
\label{tab:planning_hyperparameters}
\begin{tabular}{lc}
\toprule
Hyperparameter & Value \\
\midrule
Sampler & DDPM \\
Reverse steps & $20$ \\
Candidates & $64$ \\
Reverse-sampling temperature & $0.5$ \\
Conditioning & Current state \\
Selection score & Return energy \\
Action & First action \\
Action clipping & $[-1, 1]$ \\
\bottomrule
\end{tabular}
\end{table}

\paragraph{\algname{}-Policy.}
For \algname{}-Policy, we use the energy-guided sampler to generate synthetic trajectories for downstream offline policy optimization. The generated trajectories are annotated with the learned reward model and then filtered by domain energy and return energy. The synthetic-data construction protocol is summarized in \Cref{tab:synthetic_data_hyperparameters}. The final policy is trained with IQL on $\mathcal D_{\mathrm{syn}}\cup\mathcal D_{\mathrm{trg}}$, with architecture and hyperparameters listed in \Cref{tab:downstream_iql_hyperparameters}.

\begin{table}[t]
\centering
\small
\setlength{\tabcolsep}{7pt}
\caption{Synthetic-data construction for \algname{}-Policy.}
\label{tab:synthetic_data_hyperparameters}
\begin{tabular}{lc}
\toprule
Hyperparameter & Value \\
\midrule
Generation initialization & Gaussian noise \\
Sampler & Energy-guided reverse diffusion \\
Generation budget & $10^6$ synthetic transitions \\
Filtering order & Domain energy, then return energy \\
Filtering ratios &
$\{(0.1,0.5)\}$ \\
Reward annotation & Learned reward model $\hat r_\eta(s,a,s')$ \\
Downstream dataset & $\mathcal D_{\mathrm{syn}}\cup\mathcal D_{\mathrm{trg}}$ \\
Target data & Original $5{,}000$ target transitions \\
\bottomrule
\end{tabular}
\end{table}

\begin{table}[t]
\centering
\small
\setlength{\tabcolsep}{7pt}
\caption{Architecture and hyperparameters of downstream IQL.}
\label{tab:downstream_iql_hyperparameters}
\begin{tabular}{lc}
\toprule
Hyperparameter & Value \\
\midrule
Actor architecture & Two-layer MLP, hidden size $256$ \\
Critic architecture & Two-layer MLP, hidden size $256$ \\
Policy distribution & Tanh-squashed Gaussian \\
Log standard deviation bounds & $[-20,2]$ \\
Actor learning rate & $3\times 10^{-4}$ \\
Critic learning rate & $3\times 10^{-4}$ \\
Discount factor & $0.99$ \\
Target update rate & $0.005$ \\
Expectile parameter & $0.7$ \\
Advantage temperature & $3.0$ \\
Batch size & $256$ per replay buffer \\
Training steps & $5\times 10^5$ \\
Evaluation frequency & Every $10{,}000$ steps \\
Evaluation episodes & $10$ \\
Random seeds & $\{0,1,2\}$ \\
\bottomrule
\end{tabular}
\end{table}

\section{Additional Results}

\subsection{Results on Kinematic and Morphology Shifts}
\label{app:kinematic}

We further evaluate \algname{} on kinematic and morphology shifts in MuJoCo. Compared with gravity and friction shifts, these settings introduce more structural changes to the dynamics by modifying joint constraints or robot morphology. The results are reported in \Cref{exp:mujoco_kinematic_morphology_policy}. Across 32 tasks, \algname{} achieves the best or second-best performance in 24 tasks and obtains the highest aggregate score, showing that trajectory-level energy-guided generation remains effective under more diverse dynamics shifts.

\begin{table}[ht]
\centering
\caption{Performance comparison on MuJoCo environments with kinematic and morphology shifts. Values are normalized scores. Best and second-best are marked per row based on the mean score. \textbf{M} and \textbf{H} denote medium and hard levels.}
\resizebox{\textwidth}{!}{
\begin{tabular}{lllcccccc|cc}
\toprule
Env & Type & Level & BOSA & IQL & TD3-BC & DARA & MOBODY & Diffuser & \algname{}-Planner & \algname{}-Policy \\

\midrule
\multirow{8}{*}{HalfCheetah} & \multirow{2}{*}{morph-thigh} & M & 22.83 $\pm$ 0.03 & 20.49 $\pm$ 0.50 & 19.49 $\pm$ 0.50 & 10.90 $\pm$ 0.43 & \secondbestSc{27.18 $\pm$ 6.80} & 12.22 $\pm$ 0.12 & 12.31 $\pm$ 0.59 & \bestSc{\textbf{28.91 $\pm$ 0.41}} \\
 &  & H & 20.77 $\pm$ 0.66 & 21.69 $\pm$ 0.58 & 22.19 $\pm$ 1.08 & 10.35 $\pm$ 2.10 & \secondbestSc{28.51 $\pm$ 9.20} & 28.31 $\pm$ 0.16 & \bestSc{\textbf{29.37 $\pm$ 5.12}} & 26.19 $\pm$ 2.53 \\

 & \multirow{2}{*}{morph-torso} & M & 1.67 $\pm$ 0.87 & 1.87 $\pm$ 0.80 & 5.86 $\pm$ 0.21 & 2.91 $\pm$ 0.08 & \secondbestSc{23.92 $\pm$ 12.24} & 17.49 $\pm$ 1.63 & 15.34 $\pm$ 0.92 & \bestSc{\textbf{26.54 $\pm$ 0.98}} \\
 &  & H & 17.09 $\pm$ 15.71 & 27.81 $\pm$ 3.14 & 2.73 $\pm$ 1.25 & 29.41 $\pm$ 7.88 & \bestSc{\textbf{40.45 $\pm$ 1.26}} & 36.18 $\pm$ 4.94 & 33.08 $\pm$ 1.10 & \secondbestSc{39.98 $\pm$ 1.32} \\

 & \multirow{2}{*}{kin-footjnt} & M & \secondbestSc{36.79 $\pm$ 0.92} & 34.71 $\pm$ 0.72 & 30.19 $\pm$ 3.73 & 33.48 $\pm$ 0.34 & 31.88 $\pm$ 3.70 & 22.53 $\pm$ 3.36 & 29.85 $\pm$ 1.34 & \bestSc{\textbf{42.14 $\pm$ 3.45}} \\
 &  & H & 14.70 $\pm$ 0.92 & \bestSc{\textbf{31.68 $\pm$ 2.35}} & 14.05 $\pm$ 2.96 & \secondbestSc{31.19 $\pm$ 4.08} & 18.51 $\pm$ 7.30 & 14.69 $\pm$ 2.96 & 27.08 $\pm$ 5.41 & 12.97 $\pm$ 5.42 \\

 & \multirow{2}{*}{kin-thighjnt} & M & 14.92 $\pm$ 0.01 & 41.27 $\pm$ 3.16 & 41.77 $\pm$ 2.66 & 15.47 $\pm$ 0.62 & 59.17 $\pm$ 0.85 & \secondbestSc{61.98 $\pm$ 8.87} & 61.41 $\pm$ 4.18 & \bestSc{\textbf{62.86 $\pm$ 6.07}} \\
 &  & H & 31.72 $\pm$ 0.17 & 31.60 $\pm$ 9.36 & 31.10 $\pm$ 9.86 & 31.46 $\pm$ 2.31 & \secondbestSc{56.72 $\pm$ 0.08} & 34.49 $\pm$ 2.82 & 48.27 $\pm$ 3.42 & \bestSc{\textbf{58.43 $\pm$ 3.39}} \\

\midrule
\multirow{8}{*}{Ant} & \multirow{2}{*}{morph-halflegs} & M & 49.94 $\pm$ 5.98 & 73.65 $\pm$ 2.70 & 46.60 $\pm$ 6.24 & 70.66 $\pm$ 3.36 & \bestSc{\textbf{79.25 $\pm$ 0.61}} & 71.46 $\pm$ 4.22 & 74.29 $\pm$ 1.45 & \secondbestSc{75.08 $\pm$ 4.13} \\
 &  & H & 58.40 $\pm$ 3.41 & 57.51 $\pm$ 1.25 & 45.07 $\pm$ 2.82 & 58.46 $\pm$ 4.45 & \secondbestSc{63.76 $\pm$ 3.27} & 59.95 $\pm$ 2.51 & 53.12 $\pm$ 2.58 & \bestSc{\textbf{66.39 $\pm$ 4.59}} \\

 & \multirow{2}{*}{morph-alllegs} & M & \secondbestSc{72.02 $\pm$ 3.57} & 61.12 $\pm$ 9.73 & 47.18 $\pm$ 6.89 & 64.83 $\pm$ 4.49 & \bestSc{\textbf{75.24 $\pm$ 7.85}} & 58.01 $\pm$ 6.63 & 45.29 $\pm$ 0.47 & 59.49 $\pm$ 2.43 \\
 &  & H & 18.50 $\pm$ 4.33 & 10.44 $\pm$ 0.51 & 14.53 $\pm$ 3.74 & 4.47 $\pm$ 6.18 & \secondbestSc{24.13 $\pm$ 0.10} & 14.50 $\pm$ 0.16 & 11.94 $\pm$ 0.92 & \bestSc{\textbf{27.05 $\pm$ 5.19}} \\

 & \multirow{2}{*}{kin-anklejnt} & M & 72.06 $\pm$ 4.63 & \bestSc{\textbf{77.60 $\pm$ 3.35}} & 44.72 $\pm$ 15.96 & 75.43 $\pm$ 2.03 & 74.92 $\pm$ 6.46 & 75.51 $\pm$ 1.51 & \secondbestSc{76.43 $\pm$ 0.41} & 75.19 $\pm$ 4.62 \\
 &  & H & 63.78 $\pm$ 7.97 & 62.95 $\pm$ 7.88 & 66.22 $\pm$ 26.98 & 61.06 $\pm$ 4.92 & \bestSc{\textbf{76.97 $\pm$ 8.36}} & 52.49 $\pm$ 5.70 & 53.29 $\pm$ 1.99 & \secondbestSc{73.51 $\pm$ 5.23} \\

 & \multirow{2}{*}{kin-hipjnt} & M & 38.52 $\pm$ 5.88 & \bestSc{\textbf{60.97 $\pm$ 1.72}} & 26.85 $\pm$ 4.26 & 55.73 $\pm$ 1.93 & 54.75 $\pm$ 4.58 & 42.24 $\pm$ 3.47 & 57.21 $\pm$ 7.26 & \secondbestSc{59.07 $\pm$ 1.13} \\
 &  & H & 50.57 $\pm$ 4.89 & \secondbestSc{59.31 $\pm$ 2.92} & 33.85 $\pm$ 5.59 & 58.47 $\pm$ 3.42 & \bestSc{\textbf{59.61 $\pm$ 3.11}} & 51.08 $\pm$ 4.29 & 44.01 $\pm$ 3.58 & 54.91 $\pm$ 4.45 \\

\midrule
\multirow{8}{*}{Walker2d} & \multirow{2}{*}{morph-torso} & M & 8.26 $\pm$ 4.83 & 12.35 $\pm$ 1.45 & 18.93 $\pm$ 9.36 & 15.79 $\pm$ 1.33 & 38.67 $\pm$ 2.05 & \secondbestSc{47.97 $\pm$ 7.03} & \bestSc{\textbf{53.28 $\pm$ 7.27}} & 39.54 $\pm$ 1.50 \\
 &  & H & 1.61 $\pm$ 0.12 & 2.30 $\pm$ 0.58 & 1.54 $\pm$ 0.44 & 3.32 $\pm$ 1.13 & \bestSc{\textbf{11.96 $\pm$ 5.41}} & 1.04 $\pm$ 0.23 & 4.55 $\pm$ 0.95 & \secondbestSc{5.63 $\pm$ 2.51} \\

 & \multirow{2}{*}{morph-leg} & M & \secondbestSc{46.70 $\pm$ 8.39} & 41.12 $\pm$ 13.58 & 22.24 $\pm$ 9.95 & 39.71 $\pm$ 13.67 & \bestSc{\textbf{57.57 $\pm$ 2.00}} & 23.50 $\pm$ 3.49 & 27.09 $\pm$ 0.58 & 42.53 $\pm$ 1.51 \\
 &  & H & 14.37 $\pm$ 3.34 & 16.15 $\pm$ 3.70 & 49.07 $\pm$ 2.38 & 13.13 $\pm$ 1.24 & 49.12 $\pm$ 0.52 & \secondbestSc{49.17 $\pm$ 2.72} & 38.08 $\pm$ 1.84 & \bestSc{\textbf{51.24 $\pm$ 7.68}} \\

 & \multirow{2}{*}{kin-footjnt} & M & 17.99 $\pm$ 1.15 & 56.62 $\pm$ 12.10 & 43.31 $\pm$ 20.48 & 55.81 $\pm$ 1.36 & \bestSc{\textbf{67.56 $\pm$ 3.05}} & 52.46 $\pm$ 3.00 & 61.43 $\pm$ 5.48 & \secondbestSc{64.06 $\pm$ 0.55} \\
 &  & H & 25.76 $\pm$ 15.99 & 6.52 $\pm$ 1.61 & 26.34 $\pm$ 13.24 & 9.63 $\pm$ 0.91 & \bestSc{\textbf{57.93 $\pm$ 0.37}} & 33.93 $\pm$ 2.92 & 19.94 $\pm$ 1.96 & \secondbestSc{45.41 $\pm$ 0.67} \\

 & \multirow{2}{*}{kin-thighjnt} & M & 47.63 $\pm$ 27.26 & 61.28 $\pm$ 14.24 & 35.64 $\pm$ 11.74 & 56.28 $\pm$ 13.79 & \bestSc{\textbf{69.48 $\pm$ 4.22}} & 41.55 $\pm$ 3.08 & 52.84 $\pm$ 0.79 & \secondbestSc{64.49 $\pm$ 11.07} \\
 &  & H & 48.66 $\pm$ 14.73 & 51.66 $\pm$ 2.05 & 43.88 $\pm$ 11.54 & 63.76 $\pm$ 2.06 & \bestSc{\textbf{78.14 $\pm$ 2.50}} & 54.35 $\pm$ 4.75 & 72.19 $\pm$ 4.61 & \secondbestSc{77.89 $\pm$ 3.73} \\

\midrule
\multirow{8}{*}{Hopper} & \multirow{2}{*}{morph-foot} & M & 12.67 $\pm$ 0.00 & 32.99 $\pm$ 0.16 & 12.69 $\pm$ 0.43 & 40.61 $\pm$ 1.64 & 13.05 $\pm$ 0.48 & 40.84 $\pm$ 2.47 & \secondbestSc{41.65 $\pm$ 1.15} & \bestSc{\textbf{53.29 $\pm$ 1.68}} \\
 &  & H & 10.13 $\pm$ 0.62 & 11.78 $\pm$ 0.09 & 14.15 $\pm$ 4.30 & 13.32 $\pm$ 1.48 & \secondbestSc{65.02 $\pm$ 11.98} & 36.62 $\pm$ 2.68 & 49.67 $\pm$ 1.36 & \bestSc{\textbf{66.82 $\pm$ 5.75}} \\

 & \multirow{2}{*}{morph-torso} & M & 15.88 $\pm$ 1.18 & 13.38 $\pm$ 0.05 & 13.94 $\pm$ 0.75 & 13.29 $\pm$ 0.19 & 20.23 $\pm$ 1.29 & 22.05 $\pm$ 2.67 & \secondbestSc{22.97 $\pm$ 2.04} & \bestSc{\textbf{28.42 $\pm$ 2.72}} \\
 &  & H & 11.73 $\pm$ 0.33 & 7.77 $\pm$ 3.73 & 11.54 $\pm$ 0.81 & 4.15 $\pm$ 0.05 & 12.34 $\pm$ 0.20 & \secondbestSc{12.35 $\pm$ 1.27} & 9.41 $\pm$ 1.56 & \bestSc{\textbf{12.75 $\pm$ 0.19}} \\

 & \multirow{2}{*}{kin-legjnt} & M & 36.51 $\pm$ 1.51 & 42.28 $\pm$ 0.08 & 11.76 $\pm$ 4.60 & 44.67 $\pm$ 0.58 & 54.89 $\pm$ 0.26 & 57.56 $\pm$ 12.84 & \secondbestSc{62.01 $\pm$ 2.00} & \bestSc{\textbf{66.78 $\pm$ 11.34}} \\
 &  & H & 36.13 $\pm$ 1.70 & 45.02 $\pm$ 4.08 & 18.87 $\pm$ 1.46 & \bestSc{\textbf{65.44 $\pm$ 4.10}} & \secondbestSc{56.88 $\pm$ 3.68} & 34.97 $\pm$ 7.97 & 32.19 $\pm$ 1.41 & 42.28 $\pm$ 2.28 \\

 & \multirow{2}{*}{kin-footjnt} & M & 14.92 $\pm$ 0.01 & 15.58 $\pm$ 0.11 & 17.09 $\pm$ 0.04 & 15.47 $\pm$ 0.62 & 33.94 $\pm$ 14.81 & 27.56 $\pm$ 1.56 & \secondbestSc{55.19 $\pm$ 4.14} & \bestSc{\textbf{62.39 $\pm$ 6.01}} \\
 &  & H & 31.72 $\pm$ 0.17 & 32.41 $\pm$ 0.16 & 32.21 $\pm$ 0.00 & 32.99 $\pm$ 0.78 & 33.35 $\pm$ 0.89 & 31.29 $\pm$ 2.61 & \secondbestSc{35.93 $\pm$ 4.41} & \bestSc{\textbf{39.13 $\pm$ 1.32}} \\
 \midrule

Total &&&
964.95 
& 1123.88 
& 865.60 
& 1101.65 
& \secondbestSc{1515.10}
& 1220.34 
& 1310.71 
& \textbf{\bestSc{1551.36}} \\
\bottomrule
\end{tabular}}
\label{exp:mujoco_kinematic_morphology_policy}
\end{table}

\subsection{Additional results on energy guidance}
\label{app:only_guidance}

As shown in \Cref{tab:energy_components}, the full \algname{}-Planner achieves the best performance across HalfCheetah and Walker2d settings in this ablation. The full planner benefits from the synergy of these energy terms which leads to more stable planning performance under dynamics shift. 

\begin{table}[t]
\centering
\caption{Planner-only ablation of energy guidance on HalfCheetah and Walker2d. All variants use the same source trajectory diffusion model, planning protocol, and candidate-selection score. We report normalized target-domain scores (mean $\pm$ std over three seeds).}
\label{tab:energy_components}
\resizebox{\textwidth}{!}{
\begin{tabular}{ccccccc}
\toprule
Env & Level & No Guidance & Domain Only & Return Only & Policy Only & \algname-Planner \\
\midrule
\multirow{4}{*}{\shortstack{HalfCheetah\\Gravity}}
& 0.1 & $6.83\pm1.64$ & $8.90\pm1.91$ & $8.20\pm0.94$ & $7.12\pm0.61$ & $\mathbf{10.60}\pm\mathbf{1.84}$ \\
& 0.5 & $48.26\pm0.10$ & $47.35\pm1.51$ & $48.92\pm0.95$ & $48.63\pm1.15$ & $\mathbf{49.84}\pm\mathbf{0.17}$ \\
& 2.0 & $26.75\pm0.01$ & $27.04\pm0.05$ & $26.98\pm0.03$ & $27.05\pm0.03$ & $\mathbf{27.19}\pm\mathbf{0.06}$ \\
& 5.0 & $-0.17\pm0.04$ & $3.41\pm0.04$ & $1.24\pm0.06$ & $0.48\pm0.006$ & $\mathbf{4.04}\pm\mathbf{0.07}$ \\
\midrule
\multirow{4}{*}{\shortstack{HalfCheetah\\Friction}}
& 0.1 & $10.03\pm0.10$ & $10.32\pm0.04$ & $10.50\pm0.08$ & $10.33\pm0.07$ & $\mathbf{10.55}\pm\mathbf{0.05}$ \\
& 0.5 & $71.07\pm0.19$ & $71.74\pm0.03$ & $71.96\pm0.23$ & $71.16\pm0.68$ & $\mathbf{72.13}\pm\mathbf{0.27}$ \\
& 2.0 & $46.98\pm0.96$ & $47.52\pm0.24$ & $47.74\pm0.15$ & $47.07\pm0.34$ & $\mathbf{47.93}\pm\mathbf{0.14}$ \\
& 5.0 & $27.20\pm0.74$ & $32.49\pm0.64$ & $24.13\pm1.38$ & $26.70\pm0.37$ & $\mathbf{32.79}\pm\mathbf{0.91}$ \\
\midrule
\multirow{4}{*}{\shortstack{Walker2d\\Gravity}}
& 0.1 & $16.84\pm0.11$ & $22.27\pm1.49$ & $21.99\pm1.69$ & $16.20\pm2.02$ & $\mathbf{34.08}\pm\mathbf{0.78}$ \\
& 0.5 & $35.58\pm1.40$ & $38.76\pm1.76$ & $27.26\pm3.03$ & $24.15\pm1.43$ & $\mathbf{47.84}\pm\mathbf{1.52}$ \\
& 2.0 & $6.80\pm0.09$ & $6.87\pm0.09$ & $6.51\pm0.14$ & $5.90\pm0.06$ & $\mathbf{6.97}\pm\mathbf{0.04}$ \\
& 5.0 & $3.84\pm0.08$ & $4.12\pm0.04$ & $4.36\pm0.18$ & $3.49\pm0.33$ & $\mathbf{5.36}\pm\mathbf{0.16}$ \\
\midrule
\multirow{4}{*}{\shortstack{Walker2d\\Friction}}
& 0.1 & $8.26\pm0.26$ & $8.70\pm0.34$ & $7.96\pm0.04$ & $8.05\pm0.31$ & $\mathbf{8.71}\pm\mathbf{0.14}$ \\
& 0.5 & $68.76\pm2.04$ & $67.65\pm0.72$ & $66.74\pm3.08$ & $57.67\pm2.16$ & $\mathbf{70.36}\pm\mathbf{0.76}$ \\
& 2.0 & $51.46\pm1.83$ & $51.69\pm0.18$ & $41.87\pm1.16$ & $25.99\pm2.50$ & $\mathbf{52.66}\pm\mathbf{1.20}$ \\
& 5.0 & $4.44\pm0.03$ & $4.90\pm0.08$ & $4.61\pm0.06$ & $4.74\pm0.14$ & $\mathbf{6.19}\pm\mathbf{0.41}$ \\
\bottomrule
\end{tabular}}
\end{table}

\subsection{Additional results on filtering ratio}
\label{app:filter_ratio}

\Cref{tab:filter_control} provides the detailed results of the filtering-ratio control experiment. We apply the same filtering criterion to both energy-guided synthetic trajectories and the original source trajectories, retain the top $\{0.1, 0.2, 0.5, 1.0\}$ fraction, and train the same downstream  IQL policy with the limited target dataset. The purpose is to isolate the effect of trajectory generation from the effect of filtering. If the gain of CEDGE-Policy were caused only by filtering, filtered source data would perform comparably to filtered synthetic data at the same retention ratio. In contrast, filtered synthetic data generally yields stronger target-domain performance, showing that CEDGE-Policy benefits from generating target-compatible trajectories rather than merely filtering the original source data.

\begin{table*}[t]
\centering
\caption{Filtering-only control. We apply the same filtering criterion to both the synthetic dataset and the original source dataset, retain the top $\{0.1,0.2,0.5,1.0\}$ fraction, and train the same downstream policy. If the improvement came solely from filtering, then filtered source data should perform comparably to filtered synthetic data at matched retention ratios.}
\label{tab:filter_control}
\scriptsize
\setlength{\tabcolsep}{3pt}
\resizebox{\textwidth}{!}{
\begin{tabular}{cccccccccc}
\toprule
\multirow{2}{*}{Task} & \multirow{2}{*}{Shift}
& \multicolumn{4}{c}{Synthetic performance}
& \multicolumn{4}{c}{Source performance} \\
\cmidrule(lr){3-6} \cmidrule(lr){7-10}
& & 0.1 & 0.2 & 0.5 & 1.0
& 0.1 & 0.2 & 0.5 & 1.0 \\
\midrule

\multirow{4}{*}{\shortstack{HalfCheetah\\Gravity}}
& 0.1 & $20.04\pm3.52$ & $12.03\pm2.87$ & $11.76\pm4.93$ & $7.58\pm2.18$
      & $3.25\pm2.78$ & $0.87\pm1.44$ & $4.46\pm2.69$ & $9.62\pm4.27$ \\
& 0.5 & $2.54\pm0.59$ & $2.62\pm0.17$ & $2.31\pm0.58$ & $2.10\pm0.90$
      & $12.87\pm1.36$ & $19.60\pm3.84$ & $14.87\pm4.32$ & $44.23\pm2.93$ \\
& 2.0 & $41.66\pm5.84$ & $25.04\pm7.90$ & $39.19\pm6.05$ & $25.24\pm7.88$
      & $26.21\pm2.71$ & $28.23\pm2.56$ & $30.66\pm0.83$ & $31.34\pm1.68$ \\
& 5.0 & $75.34\pm1.63$ & $73.61\pm0.25$ & $70.94\pm1.54$ & $71.72\pm2.13$
      & $64.61\pm8.96$ & $52.14\pm22.09$ & $52.17\pm20.60$ & $44.00\pm23.13$ \\
\midrule

\multirow{4}{*}{\shortstack{HalfCheetah\\Friction}}
& 0.1 & $64.31\pm0.39$ & $63.49\pm0.15$ & $64.19\pm0.31$ & $64.22\pm0.19$
      & $25.44\pm7.60$ & $26.36\pm8.52$ & $20.99\pm3.32$ & $26.39\pm11.35$ \\
& 0.5 & $52.06\pm6.41$ & $26.20\pm6.53$ & $30.56\pm7.01$ & $34.23\pm5.68$
      & $59.02\pm4.62$ & $61.37\pm2.63$ & $66.14\pm1.89$ & $69.80\pm0.64$ \\
& 2.0 & $49.15\pm4.03$ & $38.97\pm5.97$ & $38.03\pm4.81$ & $31.69\pm7.71$
      & $26.41\pm6.57$ & $31.46\pm5.13$ & $24.99\pm7.83$ & $46.04\pm2.04$ \\
& 5.0 & $68.56\pm1.94$ & $66.04\pm2.17$ & $66.63\pm1.08$ & $67.16\pm1.77$
      & $50.35\pm7.28$ & $51.39\pm10.36$ & $45.68\pm2.94$ & $44.96\pm6.78$ \\
\bottomrule
\end{tabular}}
\end{table*}

\clearpage
\bibliographystyle{ims}
\bibliography{reference}

\end{document}